\documentclass[manuscript,screen]{acmart}
\usepackage{multirow}
\usepackage{float}
\newcommand{\tabincell}[2]{\begin{tabular}{@{}#1@{}}#2\end{tabular}}

\AtBeginDocument{%
  \providecommand\BibTeX{{%
    \normalfont B\kern-0.5em{\scshape i\kern-0.25em b}\kern-0.8em\TeX}}}

\setcopyright{acmcopyright}
\copyrightyear{2022}
\acmYear{2022}
\acmDOI{XXXXXXX.XXXXXXX}

\acmConference[Conference acronym 'XX]{Make sure to enter the correct
  conference title from your rights confirmation emai}{June 03--05,
  2018}{Woodstock, NY}
\acmPrice{15.00}
\acmISBN{978-1-4503-XXXX-X/18/06}

\begin{document}

\title{A Comprehensive Survey and Taxonomy on Single Image Dehazing Based on Deep Learning}

\author{Jie Gui}
\email{guijie@seu.edu.cn}
\affiliation{%
  \institution{School of Cyber Science and Engineering, Southeast University and Purple Mountain Laboratories}
  \city{Nanjing}
  \state{Jiangsu}
  \country{China}
  \postcode{210000}
}

\author{Xiaofeng Cong}
\affiliation{%
  \institution{School of Cyber Science and Engineering, Southeast University}
  \city{Nanjing}
  \country{China}}
\email{cxf\_svip@163.com}

\author{Yuan Cao}
\affiliation{%
  \institution{Ocean University of China }
  \city{Qingdao}
  \country{China}
}
\email{cy8661@ouc.edu.cn}

\author{Wenqi Ren}
\affiliation{%
 \institution{Institute of Information Engineering, Chinese Academy of Sciences}
 \city{Beijing}
 \country{China}}
 \email{renwenqi@iie.ac.cn}

\author{Jun Zhang}
\affiliation{%
  \institution{Anhui University}
  \city{Hefei}
  \country{China}}
  \email{wwwzhangjun@163.com}

\author{Jing Zhang}
\affiliation{%
  \institution{The University of Sydney}
  \city{Sydney}
  \country{Australia}
  }
  \email{jing.zhang1@sydney.edu.au}

\author{Jiuxin Cao}
\affiliation{%
  \institution{School of Cyber Science and Engineering, Southeast University}
  \city{Nanjing}
  \country{China}
  }

\author{Dacheng Tao}
\affiliation{%
  \institution{JD Explore Academy, China and The University of Sydney}
  \city{Sydney}
  \country{Australia}}
\email{dacheng.tao@gmail.com}

\renewcommand{\shortauthors}{Jie Gui et al.}

\begin{abstract}
With the development of convolutional neural networks, hundreds of deep learning based dehazing methods have been proposed. In this paper, we provide a comprehensive survey on supervised, semi-supervised, and unsupervised single image dehazing. We first discuss the physical model, datasets, network modules, loss functions, and evaluation metrics that are commonly used. Then, the main contributions of various dehazing algorithms are categorized and summarized. Further, quantitative and qualitative experiments of various baseline methods are carried out. Finally, the unsolved issues and challenges that can inspire the future research are pointed out. A collection of useful dehazing materials is available at \url{https://github.com/Xiaofeng-life/AwesomeDehazing}.
\end{abstract}



\maketitle

\section{Introduction}

\begin{figure}
\centering
\includegraphics[width=4cm,height=2.5cm]{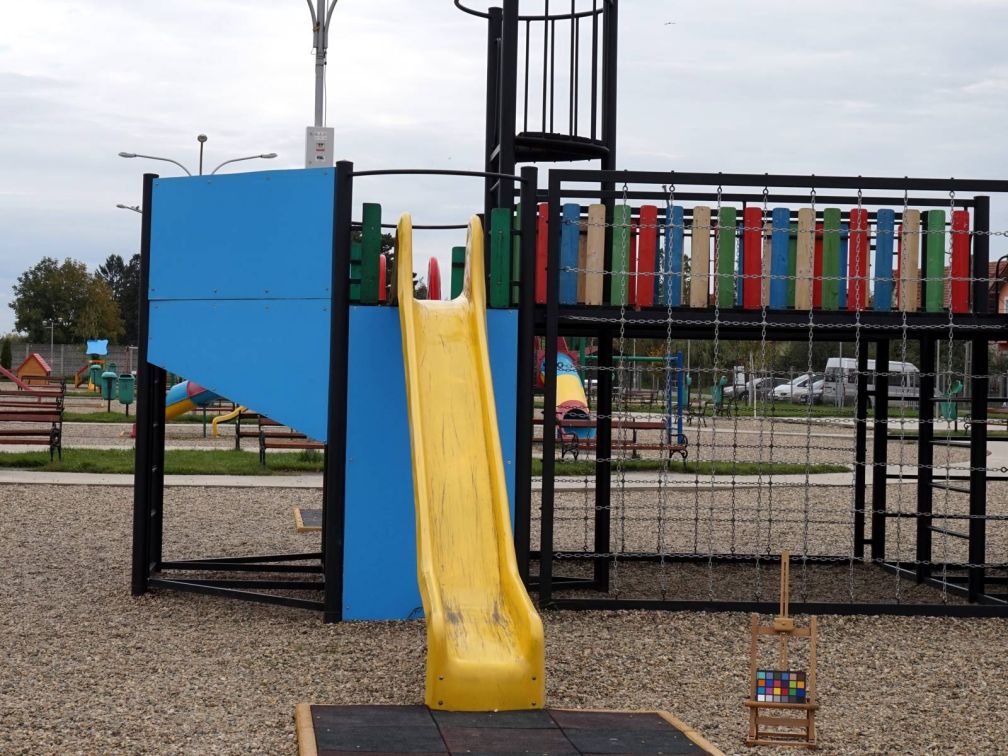}
\hspace{2cm}
\includegraphics[width=4cm,height=2.5cm]{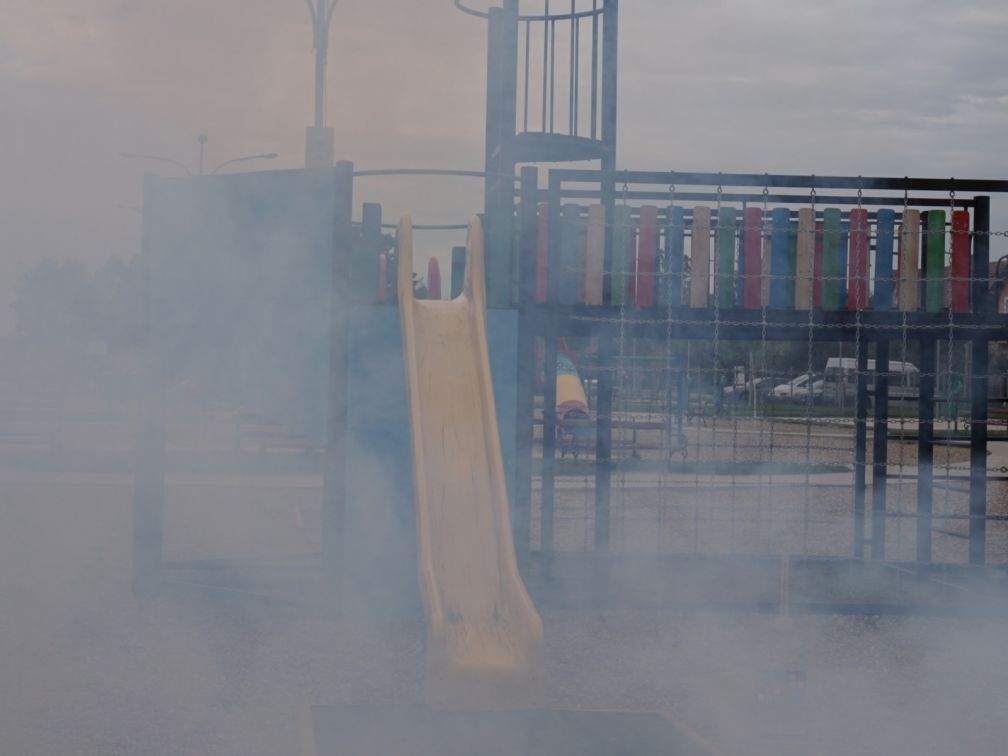}
\\

\hspace{0.5cm}(a) A clear image \hspace{3.5cm} (c) A hazy image \hspace{3.5cm}
\includegraphics[width=7cm,height=4.5cm]{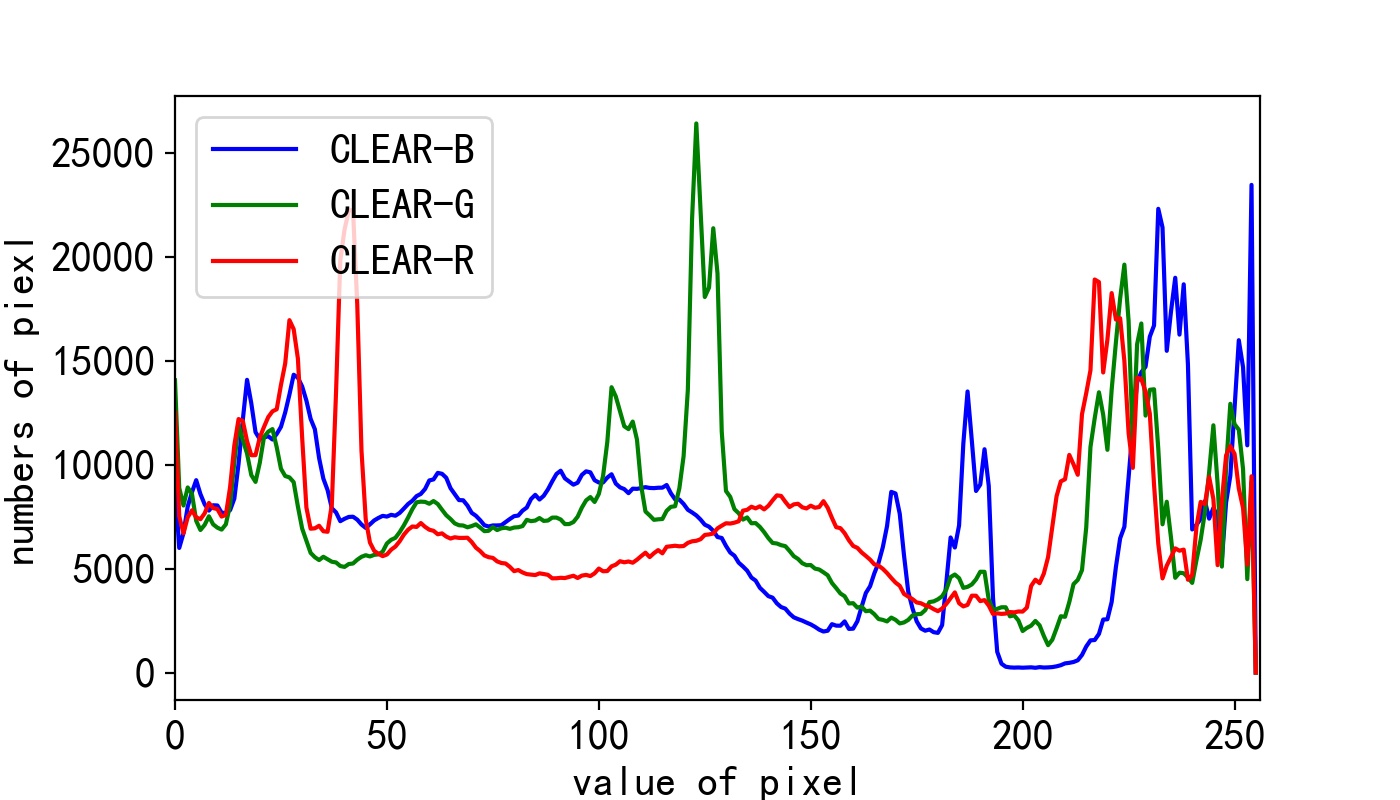}
\includegraphics[width=7cm,height=4.5cm]{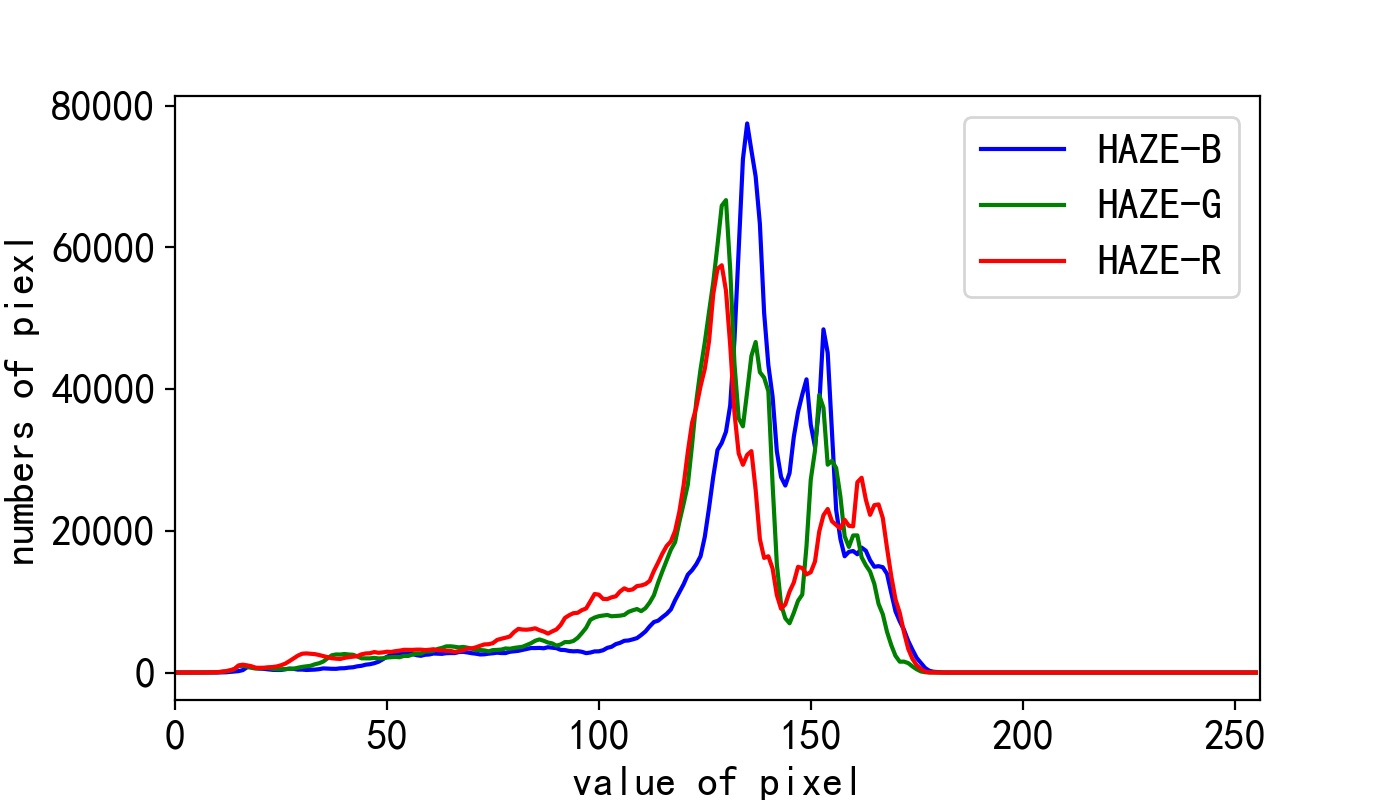}
\\

\hspace{0.5cm} (b) Histogram for (a) \hspace{2cm} (d) Histogram for (c)
\caption{Pixel histograms of clear (a) and hazy (c) images.}
  \label{fig:histogram_pixel}
\end{figure}

Due to the absorption by floating particles contained in the hazy environment, the quality of the image captured by the camera will be reduced. The phenomenon of image quality degradation in hazy weather has a negative impact on photography work. The contrast of the image will decrease and the color will shift. Meantime, the texture and edge of objects in the scene will become blurred. As shown in Fig. \ref{fig:histogram_pixel}, there is an obvious difference between the pixel histograms of hazy and haze-free images. For computer vision tasks such as object detection and image segmentation, low-quality inputs can degrade the performance of the models trained on haze-free images.

Therefore, many researchers try to recover high-quality clear scenes from hazy images. Before deep learning was widely used in computer vision tasks, image dehazing algorithms had mainly relied on various prior assumptions~\cite{he2010single} and atmospheric scattering model (ASM)~\cite{mccartney1976optics}. The processing flow of these statistical rule based methods has good interpretability. However, they may exhibit shortcomings when facing complex real world scenarios. For example, the well-known dark channel prior~\cite{he2010single} (DCP, best paper of CVPR 2009) cannot handle regions containing sky well.

Inspired by deep learning, \cite{ren2016single,cai2016dehazenet,ren2020single_he} combine ASM and convolutional neural network (CNN) to estimate the parameters of the ASM. Quantitative and qualitative experimental results show that deep learning can help the prediction of these parameters in a supervised way.

Following this, \cite{qin2020FFA,liu2019grid,liang2019selective,zhang2021hierarchical_tcsvt,zheng2021ultra} have demonstrated that end-to-end supervised dehazing networks can be implemented independently of the ASM. Thanks to the powerful feature extraction capability of CNN, these non-ASM-based dehazing algorithms can achieve comparable accuracy as ASM-based algorithms.

ASM-based and non-ASM-based supervised algorithms have shown impressive performance. However, they often require synthetic paired images that are inconsistent with real world hazy images. Therefore, recent research focus on methods that are more suitable to the real world dehazing task. \cite{cong2020discrete,alona2020un_dcp,Li2020Zero} explore unsupervised algorithms that do not require synthetic data while other studies~\cite{Li2019semi,an2021semi,chen2021psd,zhang2021single_ssdt} propose semi-supervised algorithms that exploit both synthetic paired data and real world unpaired data.

With the rapid development in this area, hundreds of dehazing methods have been proposed. To inspire and guide the future research, a comprehensive survey is urgently needed. Some papers have attempted to partially review the recent development of the dehazing research. For example, ~\cite{singh2019comprehensive,li2017haze,xu2015review} gives a summary of the non-deep learning dehazing methods, including depth estimation, wavelet, enhancement, filtering, but lacks of research on recent CNN-based methods. Parihar et al.~\cite{parihar2020comparative} provides a survey about supervised dehazing models, but it does not pay enough attention to the latest explorations of semi-supervised and unsupervised methods. Banerjee et al.~\cite{banerjee2020nighttime} introduce and group the existing nighttime image dehazing methods, however, the methods of daytime are rarely analyzed. Gui et al.~\cite{gui2021a} briefly classify and analyze supervised and unsupervised algorithms, but does not summarize the various recently proposed semi-supervised methods. Unlike existing reviews, we give a comprehensive survey on the supervised, semi-supervised and unsupervised daytime dehazing models based on deep learning.
\begin{figure}
\centering
\includegraphics[width=8cm,height=3.4cm]{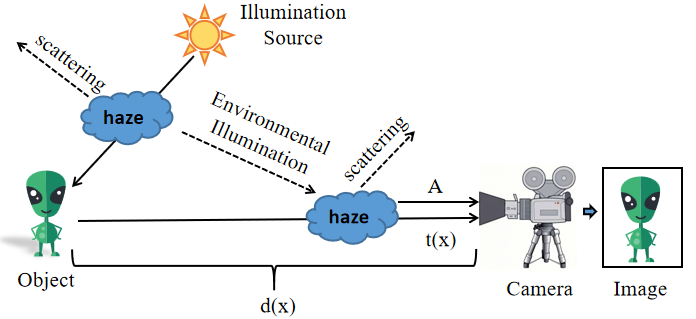}
\caption{Atmospheric Scattering Model (ASM), same as~\cite{cai2016dehazenet}.}
  \label{fig:asm}
\end{figure}

\begin{table}
\small
\centering
\caption{A taxonomy of dehazing methods. Red number index represents ASM-based methods, and black number index represents non-ASM-based methods.}
\label{tab:dehazing_methods}
\begin{tabular}{ccccc}
\hline
Category  & Key Idea & Methods \\
\hline
\multirow{17}{*}{Supervised} & Learning of $t(x)$ & \tabincell{c}{DehazeNet\textcolor{red}{~\cite{cai2016dehazenet}}, ABC-Net\textcolor{red}{~\cite{wang2020abc}}, MSCNN\textcolor{red}{~\cite{ren2016single}}, \\ MSCNN-HE\textcolor{red}{~\cite{ren2020single_he}}, SID-JMP\textcolor{red}{~\cite{huang2018single}}, LATPN\textcolor{red}{~\cite{liu2018learning}}} \\
\cline{3-3}
                                               & Joint learning of $t(x)$ and $A$&  \tabincell{c}{DCPDN\textcolor{red}{~\cite{zhang2018densely}}, DSIEN\textcolor{red}{~\cite{guo2019dense}}, LDPID\textcolor{red}{~\cite{liu2019learning}}, \\ PMHLD\textcolor{red}{~\cite{chen2020pmhld}}, HRGAN\textcolor{red}{~\cite{pang2018visual}}} \\
\cline{3-3}
                                               & Non-explicitly embedded ASM  &  \tabincell{c}{AOD-Net\textcolor{red}{~\cite{li2017all}},
                                               FAMED-Net\textcolor{red}{~\cite{zhang2020famed}},
                                               DehazeGAN\textcolor{red}{~\cite{zhu2018dehazegan}}, PFDN\textcolor{red}{~\cite{dong2020physics}}, \\ SI-DehazeGAN\textcolor{red}{~\cite{zhu2019single}}} \\
\cline{3-3}
                                               & Generative adversarial network    & \tabincell{c}{EPDN~\cite{qu2019enhanced}, PGC-UNet~\cite{zhao2020pyramid_global}, RI-GAN~\cite{dudhane2019ri}, \\ DHGAN~\cite{sim2018high}, SA-CGAN~\cite{sharma2020scale}} \\
\cline{3-3}
                                               & Level-aware         & LAP-Net\textcolor{red}{~\cite{li2019lap}}, HardGAN~\cite{deng2020hardgan} \\
\cline{3-3}
                                               & Multi-function fusion           & DMMFD\textcolor{red}{~\cite{deng2019deep}} \\
\cline{3-3}
                                               & Transformation and decomposition of input   &  GFN~\cite{ren2018gfn},  MSRL-DehazeNet~\cite{yeh2019msrl}, DPDP-Net\textcolor{red}{~\cite{yang2019dual}}, DIDH~\cite{shyam2021towards} \\
\cline{3-3}
                                               & Knowledge distillation     & KDDN~\cite{hong2020dist}, KTDN~\cite{wu2020knowledge},  SRKTDN~\cite{chen2021srktdn}, DALF~\cite{fang2021guiding} \\
\cline{3-3}
                                               & Transformation of colorspace & AIP-Net~\cite{wang2018aipnet}, MSRA-Net~\cite{sheng2022multi}, TheiaNet~\cite{mehra2021theianet}, RYF-Net\textcolor{red}{~\cite{dudhane2019ryf}}  \\
\cline{3-3}
                                               & Contrastive learning        & AECR-Net ~\cite{wu2021contrastive} \\
\cline{3-3}
                                               & Non-deterministic output   & pWAE~\cite{kim2021pixel}, DehazeFlow~\cite{li2021dehazeflow}    \\
\cline{3-3}
                                               & Retinex model                    & RDN~\cite{Li2021deep_retinex} \\
\cline{3-3}
                                               & Residual learning              & GCA-Net\cite{chen2019Gated}, DRL~\cite{du2018recursive}, SID-HL~\cite{xiao2020single}, POGAN~\cite{du2019recursive} \\
\cline{3-3}
                                               & Frequency domain              & \tabincell{c}{Wavelet U-net ~\cite{yang2019wavelet}, MsGWN~\cite{dong2020deep}, EMRA-Net~\cite{wang2021ensemble}, \\ TDN~\cite{liu2020trident}, DW-GAN~\cite{fu2021dwgan}} \\
\cline{3-3}
                                               & Joint dehazing and depth estimation        & \tabincell{c}{SDDE\textcolor{red}{~\cite{lee2020cnn}}, S2DNet\textcolor{red}{~\cite{hambarde2020s2dnet}}, DDRL\textcolor{red}{ ~\cite{guo2020reinforced}}, \\ DeAID~\cite{yang2021depth}, TSDCN-Net ~\cite{cheng2021two}} \\
\cline{3-3}
                                               & Detection and segmentation with dehazing & LEAAL\textcolor{red}{~\cite{li2020deep}}, SDNet ~\cite{zhang2021semantic}, UDnD~\cite{zhang2020unified} \\
\cline{3-3}
                                               & End-to-end CNN   & \tabincell{c}{FFA-Net~\cite{qin2020FFA}, GridDehazeNet~\cite{liu2019grid}, SAN~\cite{liang2019selective} \\  HFF~\cite{zhang2021hierarchical_tcsvt}, 4kDehazing~\cite{zheng2021ultra}} \\
\hline
\multirow{3}{*}{Semi-supervised} & Pretrain backbone and fine-tune & PSD\textcolor{red}{~\cite{chen2021psd}}, SSDT~\cite{zhang2021single_ssdt} \\
                                                         & Disentangled and reconstruction & DCNet~\cite{chen2021dcnet}, FSR\textcolor{red}{~\cite{liu2021synthetic}}, CCDM\textcolor{red}{~\cite{zhang2020color}} \\
                                                         & Two-branches training & DAID\textcolor{red}{~\cite{shao2020domain}}, SSID\textcolor{red}{~\cite{Li2019semi}}, SSIDN\textcolor{red}{~\cite{an2021semi}} \\
\hline
\multirow{4}{*}{Unsupervised}      & Unsupervised domain translation & \tabincell{c}{Cycle-Dehaze~\cite{engin2018cycle}, CDNet\textcolor{red}{~\cite{dudhane2019cdnet}}, E-CycleGAN\textcolor{red}{~\cite{liu2020end}} \\ USID\textcolor{red}{~\cite{huang2019towards}}, DCA-CycleGAN\textcolor{red}{~\cite{mo2021dca}}, DHL~\cite{cong2020discrete}} \\
\cline{3-3}
                                                         & Learning without haze-free images  & Deep-DCP\textcolor{red}{~\cite{alona2020un_dcp}} \\
                                                         & Unsupervised image decomposition & Double-DIP\textcolor{red}{~\cite{gandelsman2019double}} \\
                                                         & Zero-shot learning & ZID\textcolor{red}{~\cite{Li2020Zero}}, YOLY\textcolor{red}{~\cite{li2021yoly}} \\
\hline
\end{tabular}
\end{table}

\subsection{Scope and Goals of This Survey}
This survey does not cover all themes of dehazing research. We focus our attention on deep learning based algorithms that employ monocular daytime images. This means that we will not discuss in detail about non-deep learning dehazing, underwater dehazing~\cite{wang2021leveraging,wang2017deep,li2016underwater}, video dehazing~\cite{ren2018deep}, hyperspectral dehazing~\cite{mehta2020hidegan}, nighttime dehazing~\cite{zhang2020nighttime,zhang2017fast,zhang2014nighttime}, binocular dehazing~\cite{pang2020BidNet}, etc. Therefore, when we refer to ``dehazing'' in this paper, we usually mean deep learning based algorithms whose input data satisfies four conditions: single frame image, daytime,  monocular, on the ground. 
In summary, there are three contributions to this survey as follows.
\begin{itemize}
\item Commonly used physical models, datasets, network modules, loss functions and evaluation metrics are summarized.
\item A classification and introduction of supervised, semi-supervised, and unsupervised methods is presented.
\item According to the existing achievements and unsolved problems, the future research directions are prospected.
\end{itemize}
\subsection{A Guide for Reading This Survey}
Section \ref{sec:related_work} introduces physical model ASM, synthetic \& generated \& real world datasets, loss functions, basic modules commonly used in dehazing networks and evaluation metrics for various algorithms. Section \ref{sec:supervised} provides a comprehensive discussion of supervised dehazing algorithms. A review of semi-supervised and unsupervised dehazing methods is in Section \ref{sec:semi_supervised} and Section \ref{sec:unsupervised}, respectively. Section \ref{sec:experiment} presents quantitative and qualitative experimental results for three categories of baseline algorithms. Section \ref{sec:future} discusses the open issues of dehazing research.

A critical challenge for a logical and comprehensive review of dehazing research is how to properly classify existing methods. In the classification of Table \ref{tab:dehazing_methods}, there are several items that need to be pointed out as follows.
\begin{itemize}
\item The DCP is validated on the dehazing task, and the inference process utilizes ASM. Therefore, those methods that utilize DCP are considered to be ASM-based.

\item This survey treat the knowledge distillation based supervised dehazing network as a supervised algorithm rather than a weakly supervised/semi-supervised algorithm.

\item Supervised dehazing methods using total variation loss or GAN loss are still classified as supervised.
\end{itemize}

Here we give the notational conventions for this survey. Unless otherwise specified, all symbols have the following meanings: $I(x)$ means hazy image; $J(x)$ denotes haze-free image. $t(x)$ stands for transmission map; $x$ in $I(x)$, $J(x)$ and $t(x)$ is pixel location. The subscript ``$rec$'' denotes ``reconstructed'', such as $I_{rec}(x)$ is the reconstructed hazy image. The subscript ``$pred$'' refers to the prediction output. According to \cite{gandelsman2019double}, atmospheric light may be regarded as a constant $A$ or a non-uniform matrix $A(x)$. In this survey, atmospheric light is uniformly denoted as $A$. In addition, many papers give the proposed algorithm an abbreviated name, such as GFN~\cite{ren2018gfn} (gated fusion network for single image dehazing). For readability, this survey uses abbreviations as references to these papers. For a small amount of papers that do not give a name for their algorithms, we designate the abbreviated name according to the title of the corresponding paper.


\section{Related Work}
\label{sec:related_work}
The commonalities of dehazing algorithms are mainly reflected in four aspects. First, the modeling of the network relies on physical model ASM or is completely based on neural networks. Second, the dataset used for training the network needs to contain transmission map, atmospheric light value, or paired supervision information. Third, the dehazing networks employ different kinds of basic modules. Fourth, different kinds of loss functions are used for training. Based on these four factors, the researchers designed a variety of effective dehazing algorithms. Thus, this section introduces ASM, datasets, loss functions, and network architecture modules. In addition, how to evaluate the dehazing results is also discussed in this section.

\subsection{Modeling of the Dehazing Process}
\label{sub_sec_ASM}
Haze is a natural phenomenon that can be approximately explained by ASM. McCartney~\cite{mccartney1976optics} first proposed the basic ASM to describe the principles of haze formation. Then, Narasimhan~\cite{narasimhan2003contrast} and Nayar~\cite{nayar1999vision} extended and developed the ASM that is currently widely used. The ASM provides a reliable theoretical basis for the research of image dehazing. Its formula is
\begin{equation}
\label{eq:asm}
I(x) = J(x)t(x) + A(1-t(x)),
\end{equation}
where $x$ is the pixel location and $A$ means the global atmospheric light. In different papers, $A$ may be referred to as airlight or ambient light. For ease of understanding, $A$ is noted as atmospheric light in this survey. For the dehazing methods based on ASM, $A$ is usually unknown. $I(x)$ stands for the hazy image and $J(x)$ denotes the clear scene image. For most dehazing models, $I(x)$ is the input and $J(x)$ is the desired output. The $t(x)$ means the medium transmission map which is defined as
\begin{equation}
t(x) = e^{-\beta{d(x)}},
\end{equation}
where $\beta$ and $d(x)$ stands for the atmosphere scattering parameter and the depth of $I(x)$, respectively. Thus, the $t(x)$ is determined by $d(x)$, which can be used for the synthesis of hazy image. If the $t(x)$ and $A$ can be estimated, the haze-free image $J(x)$ can be obtained by the following formula:
\begin{equation}
J(x) = \frac{I(x) - A(1 - t(x))}{t(x)}.
\end{equation}

The imaging principle of ASM is shown in Fig. \ref{fig:asm}. It can be seen that the light reaching the camera from the object is affected by the particles in the air. Some works use ASM to describe the formation process of haze, and the parameters included in the atmospheric scattering model are solved in an explicit or implicit way. As shown in Table \ref{tab:dehazing_methods}, ASM has a profound impact on dehazing research, including supervised~\cite{ren2016single,cai2016dehazenet,li2017all,zhang2018densely}, semi-supervised~\cite{Li2019semi,liu2021synthetic,chen2021psd}, and unsupervised algorithms~\cite{li2021yoly,alona2020un_dcp}.
\subsection{Datasets for Dehazing Task}
\begin{figure}
\centering
\includegraphics[width=4.5cm,height=3cm]{make_NYU_379.jpg}
\includegraphics[width=4.5cm,height=3cm]{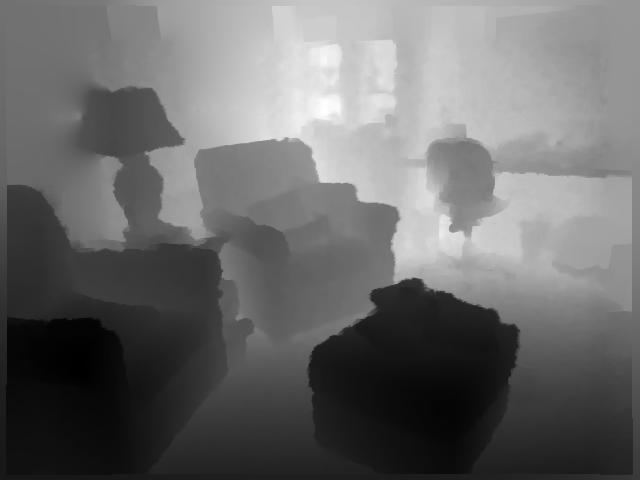}
\\
\hspace{1cm}(a) A clear image \hspace{0.5cm} (b) Depth for the clear image
\caption{A clear image and corresponding depth map in NYU-Depth dataset \cite{silberman2012indoor}.}
  \label{fig:clear_depth_nyu}
\end{figure}

\begin{figure}
\centering
\includegraphics[width=3cm,height=2cm]{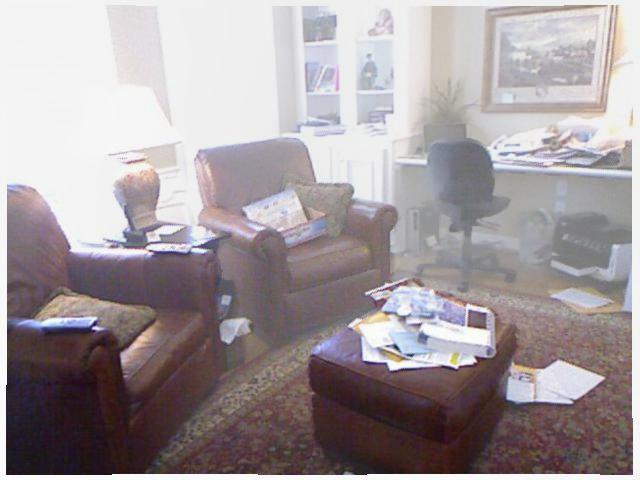}
\includegraphics[width=3cm,height=2cm]{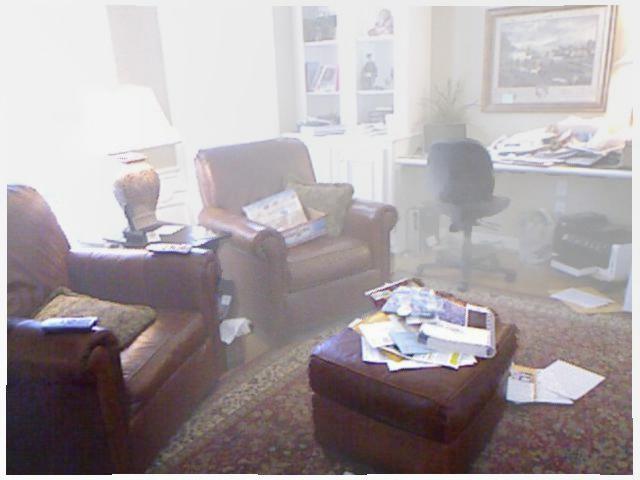}
\includegraphics[width=3cm,height=2cm]{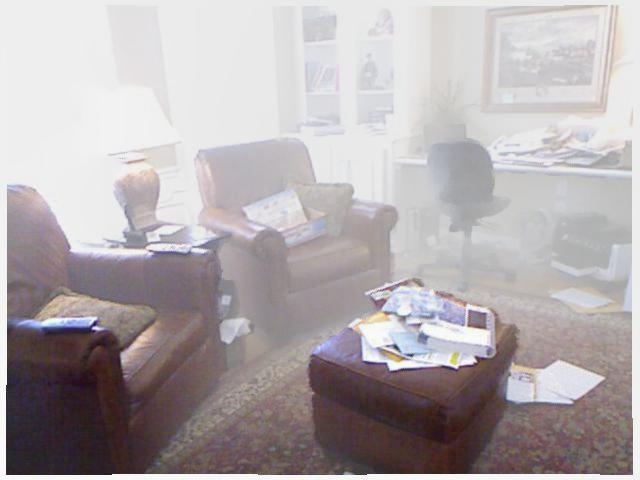}
\\
\vspace{2pt}
\includegraphics[width=3cm,height=2cm]{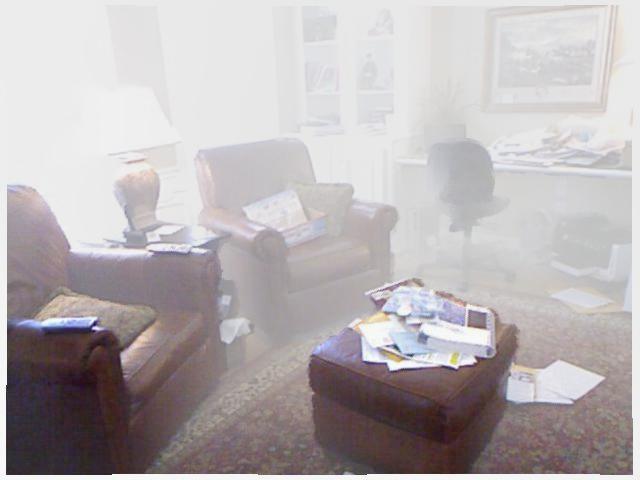}
\includegraphics[width=3cm,height=2cm]{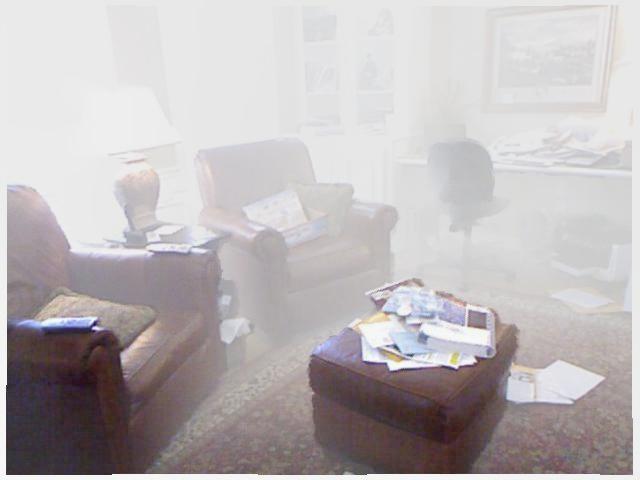}
\includegraphics[width=3cm,height=2cm]{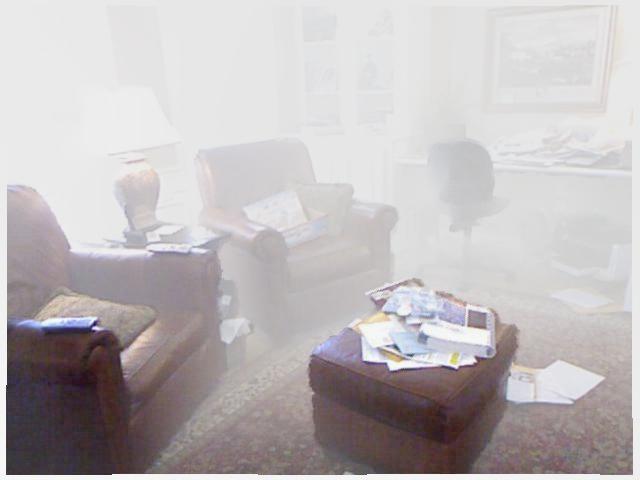}
\\
\caption{Synthesized hazy images of different densities by setting different values for the atmosphere scattering parameter $\beta$ based on NYU-Depth dataset \cite{silberman2012indoor}.}
  \label{fig:dataset_differ_haze}
\end{figure}

\begin{table}
\small
\centering
\caption{Datasets for image dehazing task. Syn stands for synthetic hazy images. HG denotes the hazy images generated from a haze generator. Real means real world scenes. S\&R denotes Syn\&Real. I and O denotes indoor and outdoor, respectively. P and NP means pair and non-pair, respectively.}
    \label{tab:dataset}
    \begin{tabular}{ccccc}
    \hline
    Dataset & type & Nums & I/O & P/NP\\
    \hline
    D-HAZY~\cite{ancuti2016d} & Syn &1400+ & I & P \\
    HazeRD~\cite{zhang2017hazeRD} & Syn & 15 & O & P\\
    I-HAZE~\cite{ancuti2018ihaze} & HG & 35 & I & P\\
    O-HAZE~\cite{ancuti2018ohaze} & HG & 45 & O & P\\
    RESIDE~\cite{Li2019bench} & S\&R &10000+ & I\&O & P\&NP\\
    Dense-Haze~\cite{anc2019den} & HG & 33 & O & P \\
    NH-HAZE~\cite{NH-Haze_2020} & HG & 55 & O & P \\
    MRFID~\cite{liu2020end} & Real & 200 & O & P\\
    BeDDE~\cite{zhao2020dehazing_eva} & Real & 200+ & O & P\\
    4KID~\cite{zheng2021ultra} & Syn & 10000 & O & P\\
    \hline
    \end{tabular}
\end{table}
For computer vision tasks such as object detection, image segmentation, and image classification, accurate ground-truth labels can be obtained with careful annotation. However, sharp, accurate and pixel-wise labels (i.e., paired haze-free images) for hazy images in natural scenes are almost impossible to obtain. Currently, there are mainly two approaches for obtaining paired hazy and haze-free images. The first way is to obtain synthetic data with the help of ASM, such as the  D-HAZY~\cite{ancuti2016d}, HazeRD~\cite{zhang2017hazeRD} and RESIDE~\cite{Li2019bench}. By selecting different parameters for ASM, researchers can easily obtain hazy images with different haze densities. Four components are needed to synthesize a hazy image: a clear image, a depth map $d(x)$ corresponding to the content of the clear image, atmospheric light $A$ and atmosphere scattering parameter $\beta$. Thus, we can divide the synthetic dataset into two stages. In the first stage, clear images and corresponding depth maps are needed to be collected in pairs. In order to ensure that the synthesized haze is as close as possible to the real world haze, the depth information must be sufficiently accurate. Fig. \ref{fig:clear_depth_nyu} shows the clear image and corresponding depth map in the NYU-Depth dataset \cite{silberman2012indoor}. In the second stage, atmospheric light $A$ and atmosphere scattering parameter $\beta$ are designated as fixed values or randomly selected. The commonly used D-HAZY~\cite{ancuti2016d} dataset is synthesized when both $A$ and $\beta$ are $1$. Several researches choose different $A$ and $\beta$ in order to increase the diversity of the synthesized images and thus improve the generalization ability of the trained model. For example, MSCNN~\cite{ren2016single} sets $A \in (0.7, 1.0)$ and $\beta \in (0.5, 1.5)$. Fig. \ref{fig:dataset_differ_haze} shows the corresponding hazy image when $\beta$ takes $6$ different values.

\begin{figure*}
\centering
\includegraphics[width=3cm,height=2cm]{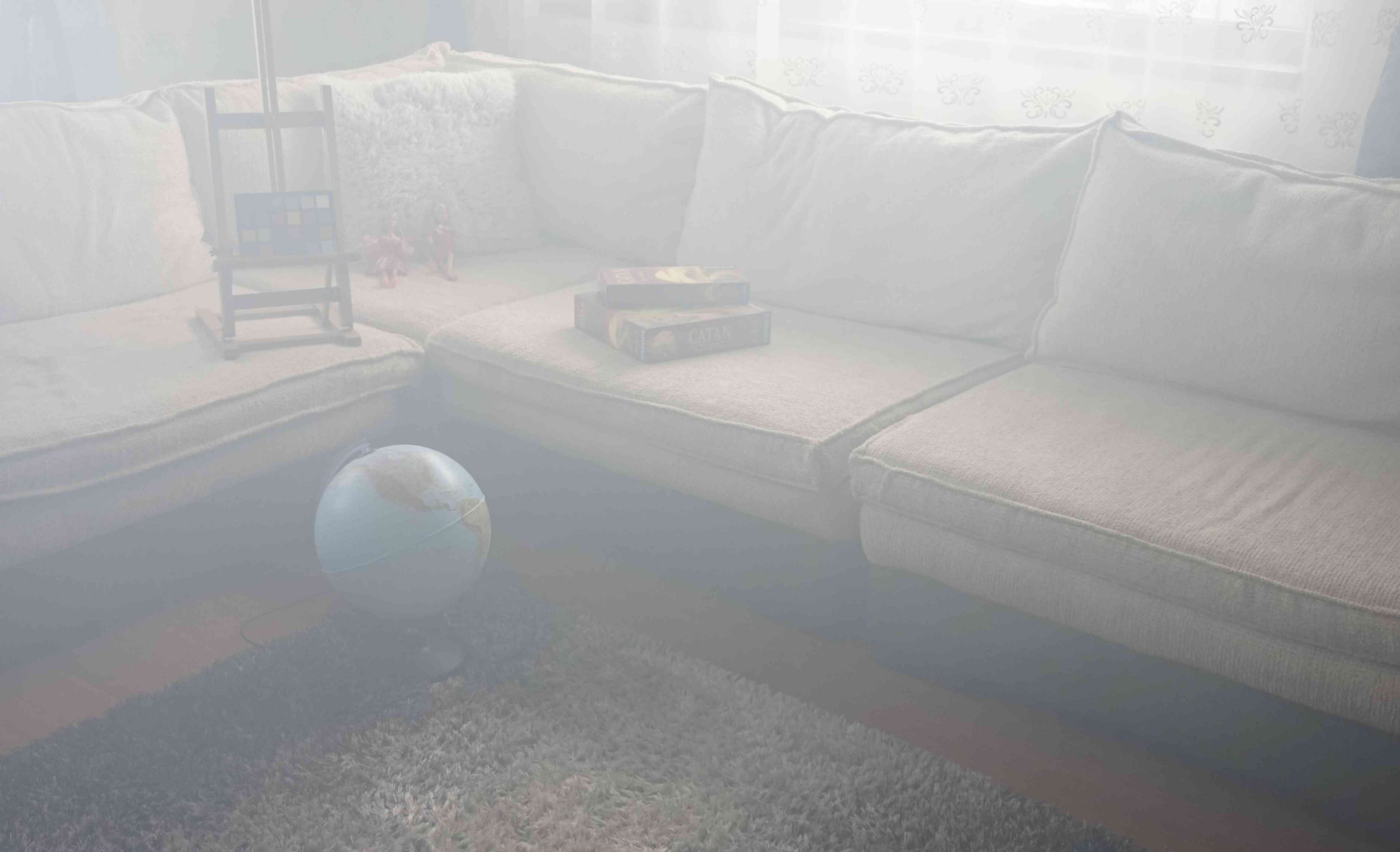}
\includegraphics[width=3cm,height=2cm]{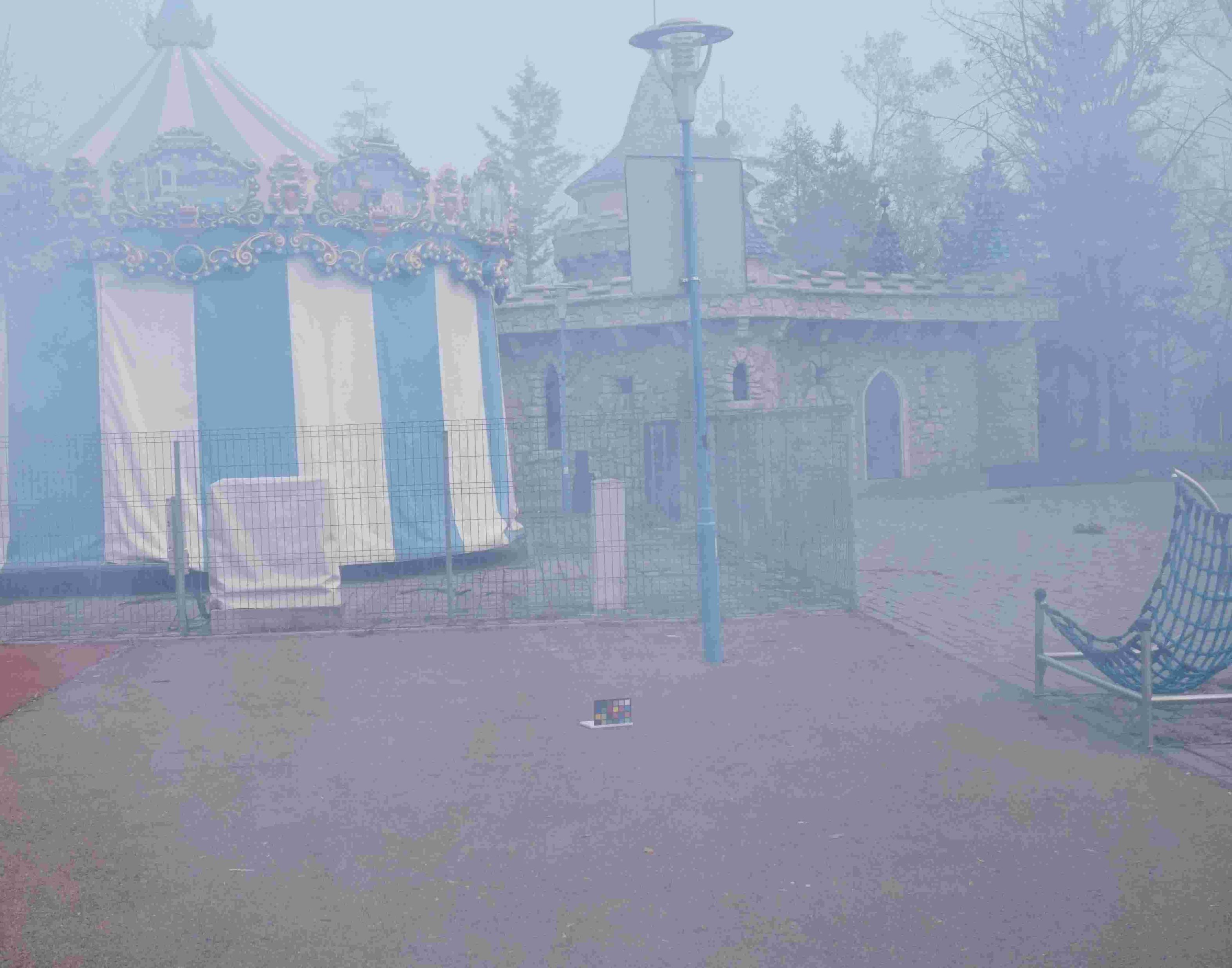}
\includegraphics[width=3cm,height=2cm]{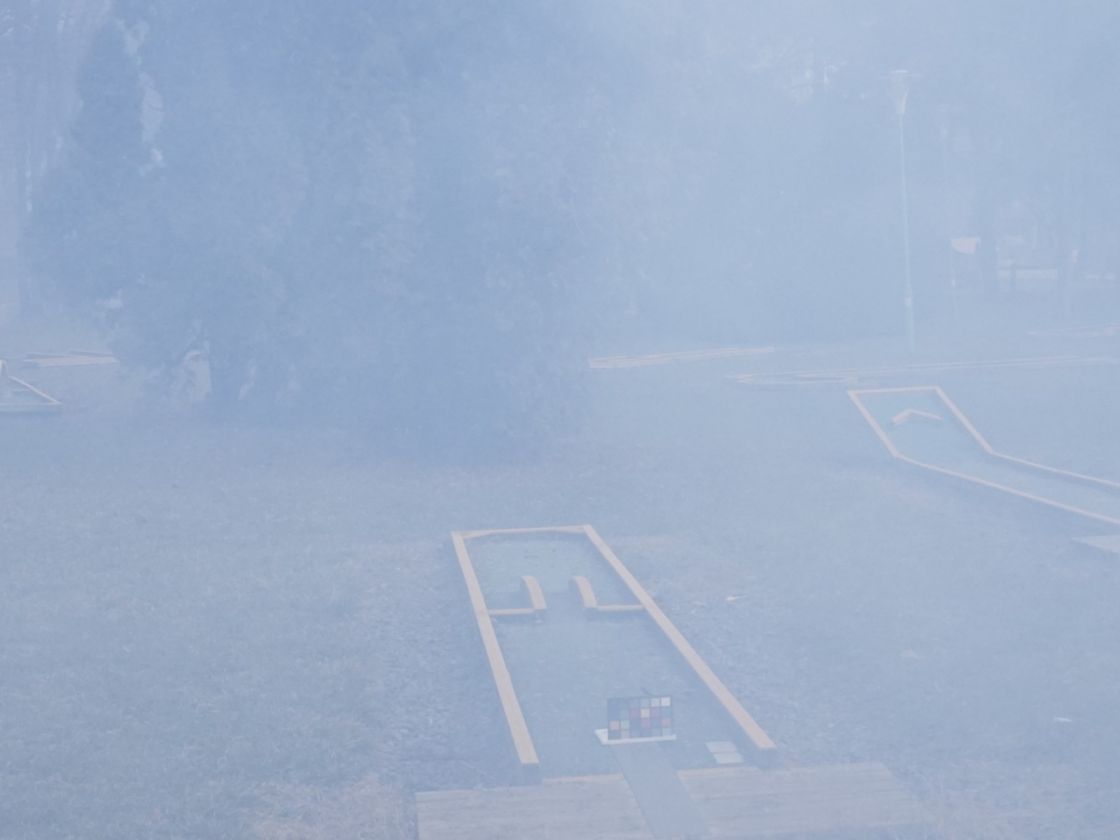}
\includegraphics[width=3cm,height=2cm]{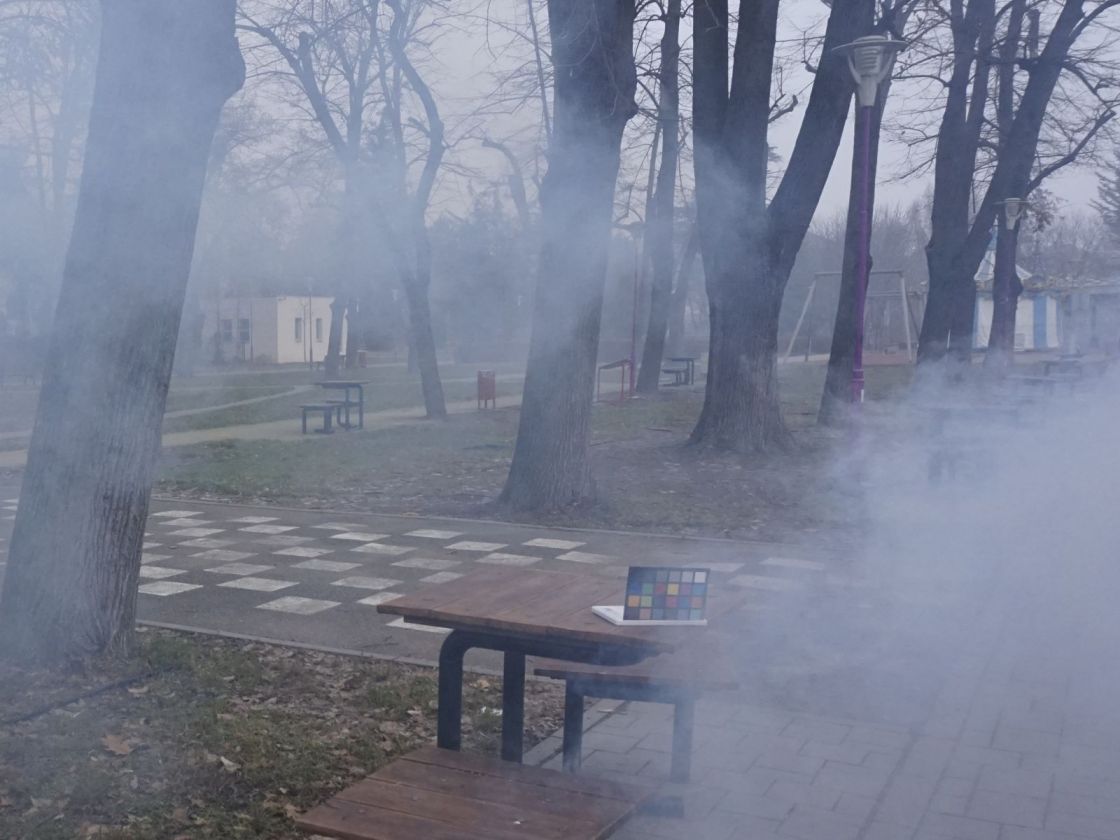}
\\
\vspace{3pt}
\includegraphics[width=3cm,height=2cm]{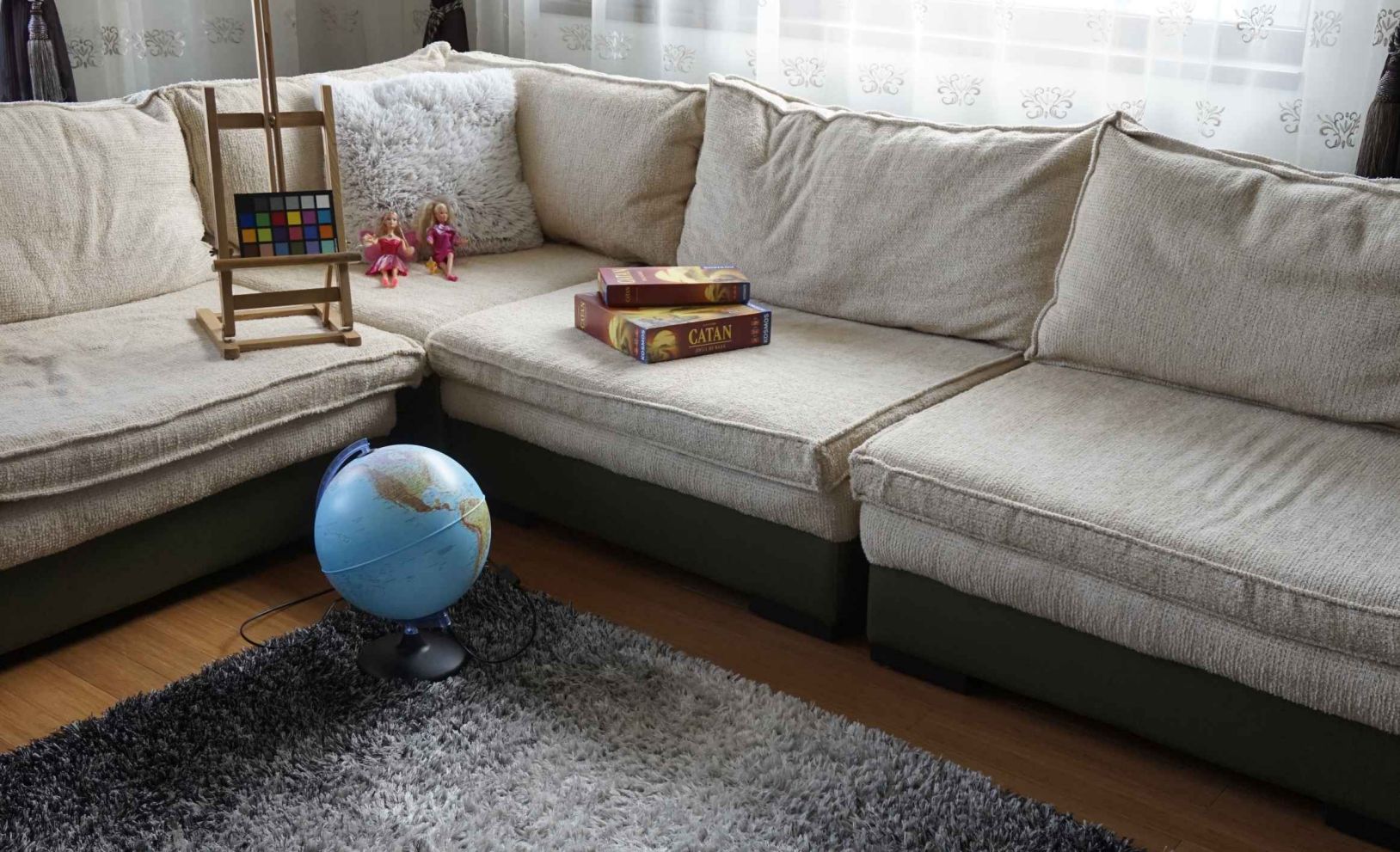}
\includegraphics[width=3cm,height=2cm]{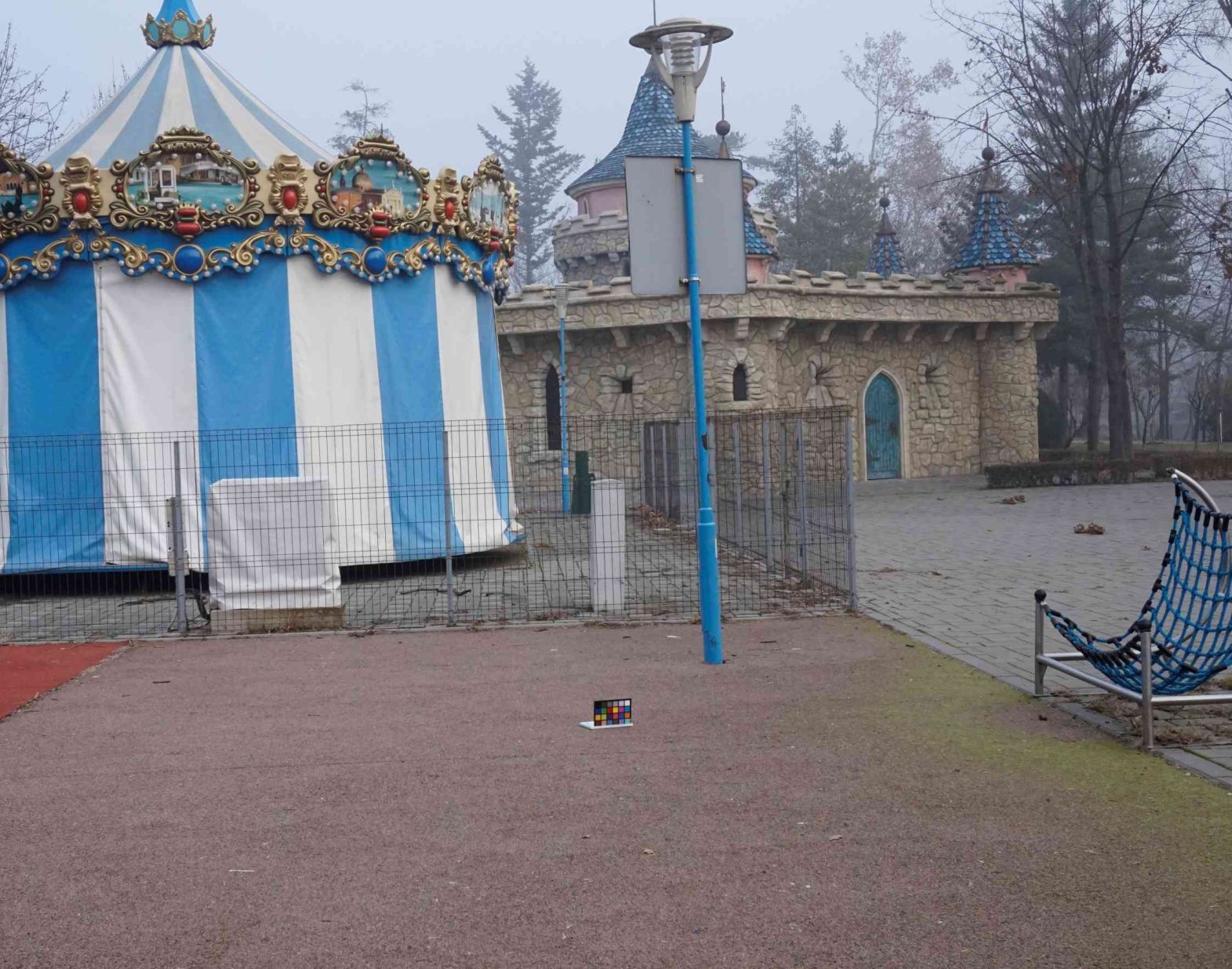}
\includegraphics[width=3cm,height=2cm]{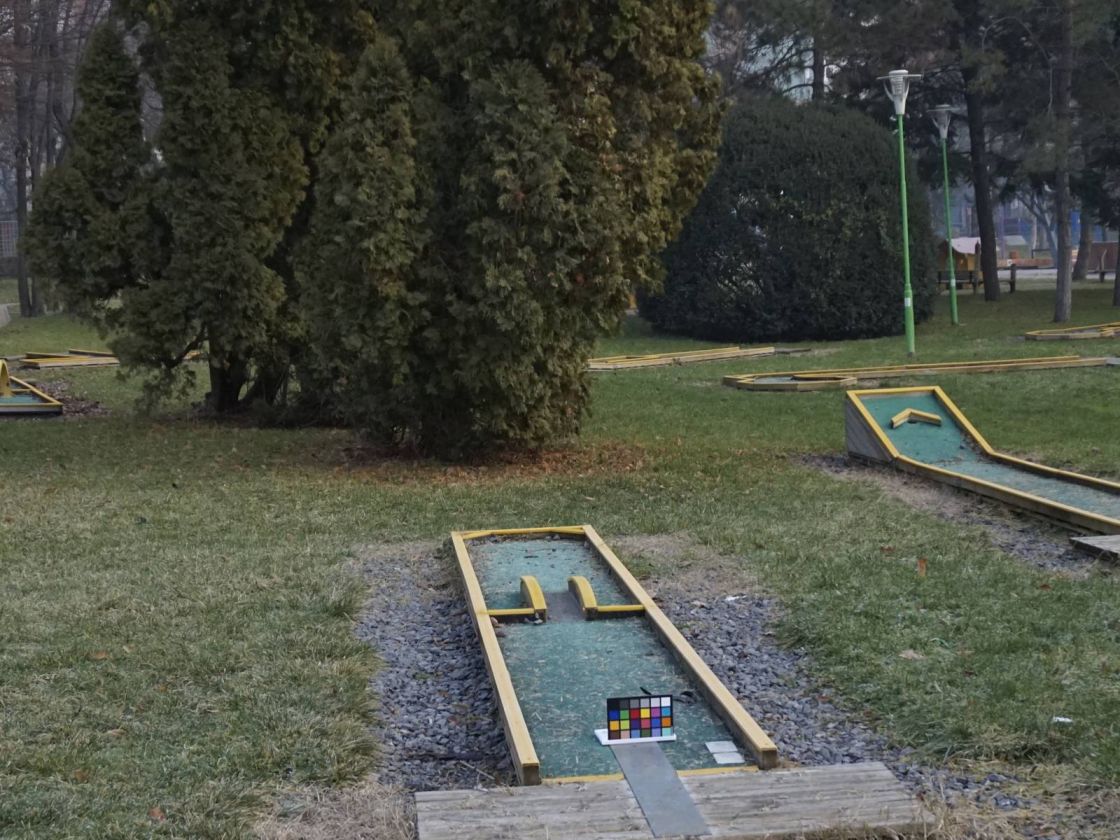}
\includegraphics[width=3cm,height=2cm]{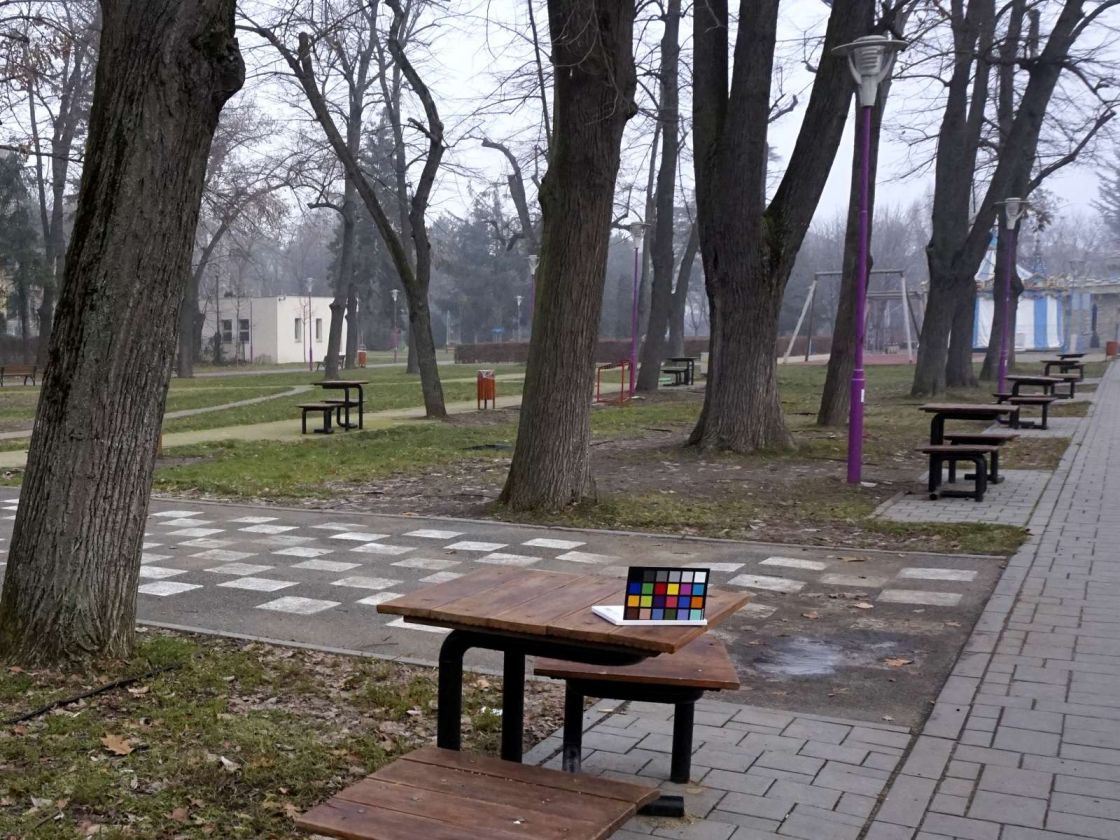}
\\
\hspace{1cm}(a) Indoor Haze \hspace{1.2cm} (b) Outdoor Haze \hspace{0.8cm}  (c) Dense Haze \hspace{0.4cm} (d) Non-Homogeneous Haze
\caption{Examples from I-HAZE~\cite{ancuti2018ihaze}, O-HAZE~\cite{ancuti2018ohaze}, Dense-Haze~\cite{anc2019den} and NH-HAZE~\cite{NH-Haze_2020}.}
  \label{fig:I_ O_Dense_NH}
\end{figure*}

The second way is to generate the hazy image by using a haze generator, such as I-HAZE~\cite{ancuti2018ihaze}, O-HAZE~\cite{ancuti2018ohaze}, Dense-Haze~\cite{anc2019den} and NH-HAZE~\cite{NH-Haze_2020}. The well-known competition New Trends in Image Restoration and Enhance (NTIRE 2018-2020) dehazing challenge ~\cite{anc2018ntire,anc2019ntire,anc2020ntire} are based on these generated datasets. Fig. \ref{fig:I_ O_Dense_NH} shows four pairs of hazy and haze-free examples contained in the datasets simulated by the haze generator. The images in Fig. \ref{fig:I_ O_Dense_NH} (a) are from indoor scenes, while (b), (c) and (d) are all pictured at outdoor views. There are differences in the pattern of haze in these three outdoor datasets. The haze in (b) and (c) is evenly distributed throughout the entire image, while the haze in (d) are non-homogeneous in the whole scenes. In addition, the density of haze in (c) is significantly higher than that in (b) and (d). These datasets with different characteristics provide useful insights for the design of dehazing algorithms. For example, in order to remove the high density of haze in Dense-Haze, it is necessary to design dehazing models with stronger feature extraction and recovery capabilities.

The main advantage of synthetic and generated haze is that it alleviates the difficulty during data acquisition. However, the hazy images synthesized based on ASM or generated by haze generator cannot perfectly simulate the formation process of real world haze. Therefore, there is an inherent difference between the synthetic and real world data. Several researches have noticed the problems of artificial data and tried to construct real world datasets, such as MRFID~\cite{liu2020end} and BeDDE~\cite{zhao2020dehazing_eva}. However, due to the high costs and difficulties of data collection, the current real world datasets do not contain enough examples as the synthetic dataset, like RESIDE~\cite{Li2019bench}. To facilitate the comparison of different datasets, we summarize the characteristics of various datasets in Table \ref{tab:dataset}.

\subsection{Network Block}
CNNs are widely used in current deep learning based dehazing networks. Commonly adopted modules are standard convolution, dilated convolution, multi-scale fusion, feature pyramid, cross-layer connection and attention. Usually, multiple basic blocks are formed into a dehazing network. In order to facilitate the understanding of the principles of different dehazing algorithms, the basic blocks commonly used in network architectures are summarized as follows.

\begin{itemize}
\item Standard convolution: It is shown that using standard convolution in a sequential connection way to build neural networks is effective. Therefore, standard convolution are often used in dehazing models~\cite{li2017all,ren2016single,sharma2020scale,zhang2020pyramid} together with other blocks.

\item Dilated convolution: Dilated convolution can increase the receptive field while keeping the size of the convolution kernel unchanged. Studies~\cite{chen2019Gated,zhang2020photo,zhang2020DRCDN,lee2020image,yan2020feature} have shown that dilated convolution can improve the performance of global feature extraction. Moreover, fusing convolution layers with different dilation rates can extract features from different receptive fields.

\item Multi-scale fusion: CNNs with multi-scale convolution kernels have been proven effective in extracting features in a variety of visual tasks~\cite{Szegedy2015going}. By using convolution kernels at different scales and fusing the extracted feature together, dehazing methods~\cite{wang2018aipnet,Tang2019single,dudhane2019ri,wang2020abc} have demonstrated that the fusion strategy can obtain the multi-scale details that are useful for image restoration. In the process of feature fusion, a common way is to spatially concatenate or add output features obtained by convolution kernels of different sizes.

\item Feature pyramid: In the research of digital image processing, the image pyramid can be used to obtain information of different resolutions. The dehazing network based on deep learning~\cite{zhang2020pyramid,zhang2018densely,zhang2018multi,singh2020single_BPP,zhao2020pyramid_global,yi2020a_novel,chen2019multi} uses this strategy in the middle layer of the network to extract multiple scales of space and channel information.

\item Cross-layer connection: In order to enhance the information exchange between different layers and improve the feature extraction ability of the network, cross-layer connections are often used in CNNs. There are mainly three types of cross-layer connections used in dehazing networks, which are residual connection~\cite{zhang2020unified,qu2019enhanced,hong2020dist,liang2019selective,chen2019sequentially} proposed by ResNet~\cite{he2016deep}, dense connection~\cite{zhu2018dehazegan,zhang2021semantic,Dong2020FD,chen2019convolutional,guo2019dense123,li2019pdr} designed by DenseNet~\cite{huang2017densely}, and skip connection~\cite{zhao2020pyramid_global,dudhane2019ri,yang2021depth,lee2020image} inspired by U-Net~\cite{ronneberger2015u}.

\item Attention in dehazing: The attention mechanism has been successfully applied in the research of natural language processing. Commonly used attention blocks in computer vision include channel attention and spatial attention. For the feature extraction and reconstruction process of 2D image, channel attention can emphasize the useful channels of the feature map. This unequal feature map processing strategy allows the model to focus more on effective feature information. The spatial attention mechanism focuses on the differences in the internal location regions of the feature map, such as the distribution of haze on the entire map. By embedding the attention module in the network, several dehazing methods~\cite{liang2019selective,chen2019sequentially,liu2019grid,qin2020FFA,yi2020a_novel,lee2020image,yan2020feature,dong2020deep,yan2020feature,zhang2020pyramid,yin2021visual,metwaly2020nonlocal,wang2021ensemble} have achieved excellent dehazing performance.
\end{itemize}

\subsection{Loss Function}
This section introduces the commonly adopted loss functions in supervised, semi-supervised and unsupervised dehazing models, which can be used for transmission map estimation, clear image prediction, hazy image reconstruction, atmospheric light regression, etc. Several algorithms use multiple losses in combination to obtain better dehazing performance. A detailed classification and summary of the loss functions used by different dehazing methods is presented in Table \ref{tab:loss_functions}. In the loss function introduced below, $X$ and $Y$ denote the predicted value and ground truth value, respectively.

\subsubsection{Fidelity Loss}
The widely used pixel-wise loss functions for dehazing research are L1 loss and L2 loss, which are defined as follows:
\begin{equation}
L1 = ||X - Y||_{1}.
\end{equation}
\begin{equation}
L2 = ||X - Y||_{2}.
\end{equation}
\subsubsection{Perceptual Loss}
The research on image super resolution \cite{ledig2017photo,wang2018esrgan} and style transfer \cite{johnson2016perceptual} indicates that the attributes of the human visual system in the process of perceptual evaluation is not fully reflected by L1 or L2 loss. Meanwhile, L2 loss may lead to over-smooth outputs \cite{ledig2017photo}. Recent researches use pre-trained classification neural networks to calculate the perceptual loss in the feature space. The most commonly used pre-training model is VGG~\cite{K2015vgg}, and some of its layers are used to calculate the distance between the predicted image and the reference image in the feature space, i.e.,
\begin{equation}
L_{per}(X, Y) = \sum_{i=1}^{N}||\psi_{i}(X) - \psi_{i}(Y)||_2,
\end{equation}
where $N$ represents the number of features selected for calculation; $i$ means the index of feature map; $\psi(\cdot)$ denotes the pretrained VGG. During the calculation of perceptual loss and network optimization, the parameters of VGG are always frozen.

\subsubsection{Structure Loss}
As a metric of the dehazing methods, Structural Similarity (SSIM)~\cite{wang2004image} is also used as a loss function in the optimization process. Studies~\cite{Dong2020FD,yu2020ensemble} have shown that SSIM loss can improve the structural similarity during image restoration. MS-SSIM~\cite{wang2003multiscale} introduces multi-scale evaluation into SSIM, which is also used as a loss function by the dehazing algorithms, i.e.,
\begin{equation}
L_{ssim} = 1 - SSIM(X, Y),
\end{equation}
\begin{equation}
L_{msssim}= 1 - MSSSIM(X, Y).
\end{equation}
\subsubsection{Gradient  Loss}
Gradient loss, also known as edge loss, is used to better restore the contour and edge information of the haze-free image. The edge extraction can be implemented as Laplacian operator, Canny operator, and so on. For example, SA-CGAN ~\cite{sharma2020scale} uses the Laplacian of Gaussian with standard deviation $\sigma$ to perform quadratic differentiation on the two-dimensional image $F$:
\begin{equation}
\begin{aligned}
L(m, n) &= \bigtriangledown^{2}F(m, n) = \frac{\partial^{2}F}{\partial{m^2}} + \frac{\partial^{2}F}{\partial{n^2}} = -\frac{1}{\pi{\sigma^{4}}}[1 - \frac{m^2 + n^2}{2\sigma^{2}}]\exp{(-\frac{m^2 + n^2}{2\sigma^{2}})},
\end{aligned}
\end{equation}
where $(m, n)$ means pixel location and $L(m, n)$ is calculated for both $X$ and $Y$, respectively. Then, regression objective functions such as L1 and L2 are used for the calculation of gradient loss.
\subsubsection{Total Variation Loss}
Total variation (TV) loss~\cite{rudin1992nonlinear} can be used to smooth image and remove noise. The training objective is to minimize the following function:
\begin{equation}
L_{TV} = ||\partial_m{X}||_1 + ||\partial_n{X}||_1,
\end{equation}
where $m$ and $n$ represent the horizontal and vertical coordinates, respectively. It can be seen from the formula that the TV loss can be added to networks trained in an unsupervised manner without using ground-truth $Y$.
\begin{table*}
\small
\centering
\caption{Loss functions for dehazing task}
\label{tab:loss_functions}
\begin{tabular}{cc}
\hline
Loss Function & Algorithms \\
\hline
L1 & ~\cite{mondal2018image,chen2019pms,deng2019deep,liang2019selective,yin2019color,dong2020physics,chen2020pmhld,yan2020feature,qin2020FFA,hong2020dist,li2020task,zhang2020nldn,zhao2020pyramid_global,park2020fusion,shin2021region,zhang2021semantic} \\
L2 & ~\cite{li2017all,pang2018visual,zhu2018dehazegan,zhang2018feed,guo2019dense,morales2019feature,chen2019sequentially,yang2019dual,Tang2019single,dong2020multi,zhang2020joint,yi2020a_novel,zhang2020multi,zhang2021single,zhang2021hierarchical_tcsvt,huang2021self,sheng2022multi} \\
SSIM & ~\cite{Dong2020FD,yu2020ensemble,metwaly2020nonlocal,wei2020single,singh2020single_BPP,li2020region,jo2021multi,shyam2021towards,zhao2020pyramid_global} \\
MS-SSIM & ~\cite{sun2021wfn,guo2019dense123,cong2020discrete,yu2021two,fu2021dwgan}\\
Perceptual & ~\cite{sim2018high,pang2018visual,zhu2019single,zhang2021hierarchical_tcsvt,wang2021lightweight,deng2020hardgan,liu2019grid,hong2020dist,qu2019enhanced,chen2019convolutional,li2019pdr,chen2019sequentially,singh2020single_BPP,Dong2020FD,shyam2021towards,engin2018cycle} \\
TV Loss & ~\cite{das2020fast,Li2019semi,shao2020domain,huang2019towards,wang2021lightweight,he2019feature}
\\
Gradient & ~\cite{zhang2018densely,zhang2019joint,zhang2020nldn,zhang2020pyramid,yi2020a_novel,zhang2021hierarchical,Li2021deep_retinex,dudhane2019ri,yin2021visual}
 \\
\hline
\end{tabular}
\end{table*}
\subsection{Image Quality Metrics}
Due to the presence of haze, the saturation and contrast of the image are reduced, and color of the image is distorted by the uncertainty mode. To measure the difference between the dehazed image and the ground truth haze-free image, objective metrics are needed to evaluate the results obtained by various dehazing algorithms.

Most papers use Peak Signal-to-Noise Ratio (PSNR)~\cite{huynh2008scope} and SSIM~\cite{wang2004image} to evaluate the image quality after dehazing. The computation of PSNR needs to use the formula (\ref{eq:mse}) to obtain the mean square error (MSE):
\begin{equation}
\label{eq:mse}
MSE = \frac{1}{H\times{W}}\sum_{i=1}^{H}\sum_{j=1}^{W}(X(i,j) - Y(i,j))^{2},
\end{equation}
where $X$ and $Y$ respectively represent two images to be evaluated. $H$ and $W$ are their height and width, that is, the dimensionalities of $X$ and $Y$ should be strictly the same. The pixel position index of the image is represented by $i$ and $j$. Then, the PSNR can be obtained by logarithmic calculation as following:
\begin{equation}
PSNR = 10log_{10}[\frac{({2^{N}-1)^2}}{MSE}],
\end{equation}
where $N$ equals $8$ for $8$-bit images. SSIM is based on the correlation between human visual perception and structural information, and its formula is defined as following:
\begin{equation}
SSIM(X,Y) = \frac{{2u_x}{u_y}+C_1}{u_x^2+u_y^2+C_1}\frac{2\sigma_{xy}+C_2}{\sigma_x^2+\sigma_y^2+C_2},
\end{equation}
where $\mu_x$, $\mu_y$, $\sigma_x$ and $\sigma_y$ represent the mean and variance of $X$ and $Y$, respectively; $\sigma_{xy}$ is the covariance between two variables; $C_1$ and $C_2$ are constants used to ensure numerical stability.

Since haze can cause the color of scenes and objects to change, some works use CIEDE2000~\cite{sharma2005ciede2000} as an assessment of the degree of color shift. PSNR, SSIM and CIEDE belong to full-reference evaluation metrics, which means that a clear image corresponding to a hazy image must be used as a reference. However, real world pairs of hazy and haze-free images are difficult to keep exactly the same in content. For example, the lighting and objects in the scene may change before and after the haze appears. Therefore, in order to maintain the accuracy of the evaluation process, it is necessary to synthesize the corresponding hazy image with a clear image~\cite{min2019quality}.

In addition to the full-reference metric PSNR, SSIM and CIEDE, the recent works~\cite{li2020deep,Li2019bench} utilizes the no-reference metrics SSEQ~\cite{liu2014no} and BLIINDS-II~\cite{saad2012blind} to evaluate dehazed images without ground truth. No-reference metric is of crucial value for real world dehazing evaluation. Nevertheless, the evaluation of current dehazing algorithms are usually conducted on datasets with pairs of hazy and haze-free images. Since the full-reference metric is more suitable for paired datasets, it is more widely used than the no-reference metric.

\section{Supervised Dehazing}
\label{sec:supervised}
Supervised dehazing models usually require different types of supervisory signals to guide the training process, such as transmission map, atmospheric light, haze-free image label, etc. Conceptually, supervised dehazing methods can be divided into ASM-based and non-ASM-based ones. However, there may be overlaps in this way, since both ASM-based and non-ASM-based algorithms may entangle with other computer vision tasks such as segmentation, detection, and depth estimation. Therefore, this section categorizes supervised algorithms according to their main contributions, so that the techniques that prove valuable for dehazing research are clearly observed.

\subsection{Learning of $t(x)$}
According to the ASM, the dehazing process can be divided into three parts: transmission map estimation, atmospheric light prediction, and haze-free image recovery. MSCNN~\cite{ren2016single} proposes the following three steps for solving ASM: (1) use CNN to estimate the transmission map $t(x)$, (2) adopt statistical rules to predict atmospheric light $A$, and (3) solve $J(x)$ by $t(x)$ and $A$ jointly. MSCNN adopts a multi-scale convolutional model for transmission map estimation and optimizes it with L2 loss. In addition, $A$ can be obtained by selecting $0.1\%$ darkest pixels in $t(x)$ corresponding the one with the highest intensity in $I(x)$~\cite{he2010single}. Thus the clear image $J(x)$ can be obtained by
\begin{equation}
J(x) = \frac{I(x) - A}{max{\{0.1,t(x)\}}} + A.
\end{equation}

Different papers may use different statistical priors to estimate $A$, but the strategies they use for dehazing are similar to MSCNN. ABC-Net~\cite{wang2020abc} uses the max pooling operation to obtain maximum value from each channel of $I(x)$. SID-JPM~\cite{huang2018single} filters each channel of a RGB input image by a minimum filter kernel. Then the maximum value of each channel is used as the estimated $A$. LAPTN~\cite{liu2018learning} also applies the minimum filter together with the maximum filter for the prediction of $A$. These methods generally do not require atmospheric light annotations, but need paired of hazy images and transmission maps.

\subsection{Joint Learning of $t(x)$ and $A$}
Instead of using convolutional networks and statistical priors jointly to estimate the physical parameters of ASM, some works implement the prediction of physical parameters entirely through CNN. DCPDN~\cite{zhang2018densely} estimates the transmission map through a pyramid densely connected encoder-decoder network, and uses a symmetric U-Net to predict the atmospheric light $A$. In order to improve the edge accuracy of the transmission map, DCPDN has designed a hybrid edge-preserving loss, which includes L2 loss, two-directional gradient loss, and feature edge loss.

DHD-Net~\cite{xie2020dhd} designs a segmentation-based haze density estimation algorithm, which can segment dense haze areas and divide the global atmosphere light $A$ candidate areas. HRGAN~\cite{pang2018visual} utilizes a multi-scale fused dilated convolutional network to predict $t(x)$, and employs a single-layer convolutional model to estimate $A$. PMHLD~\cite{chen2020pmhld} uses a patch map generator and refine the network for transmission map estimation, and utilizes VGG-16 for atmospheric light estimation. It is worth noting that if using regression training to obtain $A$, the ground truth labels are generally required.

\subsection{Non-explicitly Embedded ASM}
ASM can be incorporated into CNN in a reformulated or embedded way. AOD-Net~\cite{li2017all} founds that the end-to-end neural network can still be used to solve the ASM without directly using ground truth $t(x)$ and $A$. According to the original ASM, the expression of $J(x)$ is
\begin{equation}
J(x) = \frac{1}{t(x)}I(x) - A\frac{1}{t(x)} + A.
\end{equation}

AOD-Net proposes $K(x)$, which has no actual physical meaning, as an intermediate parameter describing $t(x)$ and $A$. $K(x)$ is defined as
\begin{equation}
K(x) = \frac{\frac{1}{t(x)}(I(x) - A) + (A - b)}{I(x) - 1},
\end{equation}
where $b$ equals $1$. According to the ASM theory, $J(x)$ can be uniquely determined by $K(x)$ as
\begin{equation}
J(x) = K(x)I(x) - K(x) + 1.
\end{equation}

AOD-Net considers that the non-joint transmission map and atmospheric light prediction process may produce accumulated errors. Therefore, independent estimation of $K(x)$ can reduce the systematic error. FAMED-Net \cite{zhang2020famed} extends this formulation in a multi-scale framework and utilizes fully point-wise convolutions to achieve fast and accurate dehazing performance. DehazeGAN~\cite{zhu2018dehazegan} incorporates the idea of differentiable programming into the estimation process of $A$ and $t(x)$. Combined with a reformulated ASM, DehazeGAN also implements an end-to-end dehazing pipeline. PFDN~\cite{dong2020physics} embeds ASM in the network design and proposes a  feature dehazing unit, which removes haze in a well-designed feature space rather in the raw image space. It is instructive that ASM can still help the dehazing task for non-explicit parameter estimation.

\begin{figure}
\centering
\includegraphics[width=8cm,height=1.78cm]{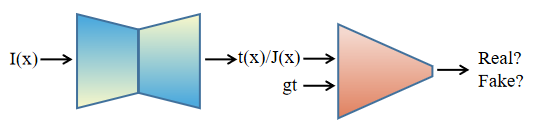}
\caption{Generative adversarial network for dehazing, where gt means ground truth.}
  \label{fig:GAN}
\end{figure}

\subsection{Generative Adversarial Networks}
Generative adversarial networks have an important impact on dehazing research. In general, supervised dehazing networks that rely on paired data can use adversarial loss as an auxiliary supervisory signal. Adversarial loss~\cite{gui2021review} can be seen as two parts: the training objective of the generator is to generate images that the discriminator considers to be real. The optimization purpose of the discriminator is to distinguish the generated image from the real image contained in the dataset as possible as it could. For dehazing task, the effect of the adversarial loss is to make the generated image closer to the real one, which is beneficial for the optimization of the haze-free $J(x)$ and transmission map $t(x)$~\cite{zhang2019joint}, as shown in Fig. \ref{fig:GAN}.

Inspired by patchGAN~\cite{isola2017image}, which can better preserve high-frequency information, DH-GAN~\cite{sim2018high}, RI-GAN~\cite{dudhane2019ri} and DehazingGAN~\cite{zhang2020photo} use $N \times N$ patches instead of a single value as the output of the discriminator. Several works explore joint training mechanisms of multiple discriminators, such as EPDN~\cite{qu2019enhanced} and PGC-UNet~\cite{zhao2020pyramid_global}. Discriminator $D_1$ is used to guide the generator on a fine scale, while discriminator $D_2$ helps the generator to produce a global realistic output on a coarse scale. In order to realize the joint training of the two discriminators, EPDN downsamples the input image of $D_1$ by a factor of $2$ as the input of $D_2$. The adversarial loss is
\begin{equation}
L_{adv} = \min_{G}[\max_{D_{1}, D_{2}} \sum_{k=1, 2}\ell_{A}(G, D_{k})],
\end{equation}
where the form of adversarial loss $\ell_{A}$ is the same as the single GAN. The exploration of GAN brings plug-and-play tools to supervised algorithms. Since the training of the discriminator can be done in an unsupervised manner, the quality of the dehazed images can be improved without the requirement of extra labels. However, the training of GAN sometimes suffers from instability and non-convergence, which may bring certain additional difficulties to the training of the dehazing network.
\subsection{Level-aware}
According to the scattering theory, the farther the scene is from the camera, the more aerosol particles pass through. This means that areas within a single hazy image that are farther from the camera have higher densities of haze. Therefore, LAP-Net~\cite{li2019lap} proposes that the algorithm should consider the difference in haze density inside the image. Through multi-stage joint training, LAP-Net implements an easy-to-hard model that focuses on a specific haze density level by a stage-wise loss
\begin{equation}
\hat{t}^{s}(x) = \begin{cases} \mathcal{F}(I(x), \theta^{s}), \text{if } s =1,  \\
\mathcal{F}(I(x), \theta^{s}, \hat{t}^{s-1}(x)), \text{if } s > 1,
\end{cases}
\end{equation}
where $\mathcal{F}$ represents the transmission map prediction network with parameter $\theta^{s}$ in stage $s$. In the first stage, the transmission map prediction network is responsible for estimating the case with mild haze. In the second and subsequent stages, the prediction result of the previous stage and the hazy image are used as joint input for processing higher haze density.

The density of haze may be related to conditions such as temperature, wind, altitude, and humidity. Thus, the formation of haze should be space-variant and non-homogeneous. Based on this observation, HardGAN~\cite{deng2020hardgan} argues that estimating the transmission map for dehazing may be inaccurate. By encoding the atmospheric brightness as $1 \times 1 \times 2$ matrix $\gamma_{i}^{G} \& \beta_{i}^{G}$ and pixel-wise spatial information as $H \times W \times 2$ matrix $\gamma_{i}^{L} \& \beta_{i}^{L}$ for $i$-th channel of the input $x$, HardGAN designs the control function of atmospheric brightness $G_{i}$ and spatial information $L_{i}$ as
\begin{gather}
G_{i} = \gamma_{i}^{G}\frac{x - \mu}{\sigma} + \beta_{i}^{G}, \notag \\
L_{i} = \gamma_{i}^{L}\frac{x - \mu}{\sigma} + \beta_{i}^{L},
\end{gather}
where $\mu$ and $\sigma$ denote the mean and standard deviation for $x$, respectively. After obtaining $G_{i}$ and $L_{i}$, HardGAN uses a linear model to fuse them for recovering haze-free image
\begin{equation}
    J_{pred_{i}}(x) = (1 - HA_{i}) * G_{i} + HA_{i} * L_{i},
\end{equation}
where $HA$ is calculated by the intermediate feature map of HardGAN via using instance normalization followed by a sigmoid layer, and $*$ denotes element-wise product.

\subsection{Multi-function Fusion}
DMMFD~\cite{deng2019deep} designs a layer separation and fusion model for improving learning ability, including reformulated ASM, multiplication, addition, exponentiation and logarithmic decomposition:
\begin{gather}
J_{0}(x) = \frac{I(x) - A_{0} \times (1 - t_{0}(x))}{t_{0}(x)}, \notag \\
J_{1}(x) = I(x) \times R_{1}(x), \notag \\
J_{2}(x) = I(x) + R_{2}(x), \notag \\
J_{3}(x) = (I(x))^{R_{3}(x)}, \notag \\
J_{4}(x) = \log(1 + I(x) \times R_{4}(x)),
\end{gather}
where $R_i(x)$ stands for layers in the network; $A_0$ and $t_{0}(x)$ are the atmospheric light and transmission map estimated by the feature extraction network, respectively. $J_{1}{(x)}$, $J_{2}{(x)}$, $J_{3}{(x)}$, and $J_{4}{(x)}$ can be used as four independent haze-layer separation models. It is based on the assumption that the input hazy image $I(x)$ can be separated into a haze-free layer $J(x)$ and another layer $H(x)$, denoted as $I(x) = \phi(J(x), H(x))$. The final dehazing result is obtained by weighted fusion of the intermediate outputs $J_{0}{(x)}$, $J_{1}{(x)}$, $J_{2}{(x)}$, $J_{3}{(x)}$ and $J_{4}{(x)}$ by five learned attention maps $W_0$, $W_1$, $W_2$, $W_3$ and $W_4$ as
\begin{equation}
\begin{aligned}
J_{pred}(x) = & W_{0} \times J_{0}{(x)} + W_{1} \times J_{1}{(x)} + W_{2} \times J_{2}{(x)} + W_{3} \times J_{3}{(x)} + W_{4} \times J_{4}{(x)}.
\end{aligned}
\end{equation}

Through ablation studies, DMMFD has demonstrated that the fusion of multiple layers can improve the quality of the scene restoration process.

\subsection{Transformation and Decomposition of Input}
GFN~\cite{ren2018gfn} proposes two observations on the influence of haze. First, under the influence of atmospheric light, the color of a hazy image may be distorted to some extent. Second, due to the existence of scattering and attenuation phenomena, the visibility of objects far away from the camera in the scene will be reduced. Therefore, GFN uses three enhancement strategies to process the original hazy image and use them as inputs to the dehazing network together. The white balanced input $I_{wb}(x)$ is obtained from the gray world assumption. The contrast enhanced input $I_{ce}(x)$ is composed of average luminance value $\widetilde{I}(x)$ and the control factor $\mu$, as following:
\begin{equation}
I_{ce}{(x)} = \mu(I(x) - \widetilde{I}{(x)}),
\end{equation}
where $\mu = 2 \cdot (0.5 + \widetilde{I}(x))$. By using the the nonlinear gamma correction, the $I_{gc}{(x)}$ used to enhance the visibility of the $I(x)$ can be obtained by
\begin{equation}
I_{gc}{(x)} = \alpha{I(x)^{\gamma}},
\end{equation}
where $\alpha = 1$, and $\gamma = 2.5$. The final dehazing image $J(x)$ is determined by the combination of three inputs, where $C_{wb}$, $C_{ce}$ and $C_{gc}$ are confidences map for fusion process:
\begin{equation}
J(x) = C_{wb} \circ {I_{wb}{(x)}} + C_{ce} \circ {I_{ce}{(x)}} + C_{gc} \circ {I_{gc}{(x)}}.
\end{equation}

MSRL-DehazeNet~\cite{yeh2019msrl} decomposes the hazy image into low frequency and high frequency as the base component $I_{base}{(x)}$ and the detail component $I_{detail}{(x)}$, respectively. The basic component can be thought as the main content of the image, while the high frequency component denotes the edge and texture. Therefore, the dehazed image can be obtained by the dehazing function $D(\cdot)$ of the basic component and the enhancement function $E(\cdot)$ of the detail component. The whole process can be represented by a linear model as following:
\begin{gather}
I(x) = I_{base}{(x)} + I_{detail}{(x)}, \notag \\
J(x) = D(I_{base}{(x)}) + E(I_{detail}{(x)}).
\end{gather}

DIDH~\cite{shyam2021towards} decomposes both the input hazy image $I(x)$ and the predicted haze-free image $J(x)$ to obtain $(I(x), LF(I(x)))$, $(I(x), HF(I(x)))$, $(J(x), LF(J(x)))$ and $J(x), HF(J(x))$, where $LF(\cdot)$ and $HF(\cdot)$ denotes Gaussian and Laplacian filter, respectively. By fusing the decomposed data with the pre-decomposition data as the input of the discriminator, DIDH can improve the quality of the image generated by the adversarial training process. With the help of the discriminators $D_{LF}$ and $D_{HF}$, the dehazing network can be optimized in an adversarial way.

Compared with conventional data augmentation, such as rotation, horizontal or vertical mirror symmetry, and random cropping, the transformation and decomposition of the input is a more efficient strategy for the usage of hazy images.
\begin{figure}
\centering
\includegraphics[width=6cm,height=2.9cm]{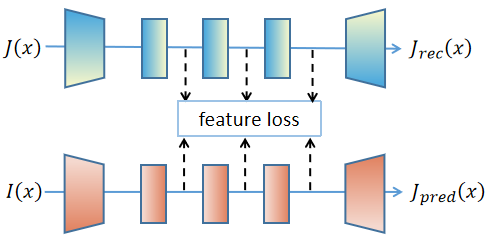}
\\
\caption{A general knowledge distillation strategy used in KDDN~\cite{hong2020dist}, KTDN~\cite{wu2020knowledge}, SRKTDN~\cite{chen2021srktdn}, etc.}
  \label{fig:kddn}
\end{figure}

\subsection{Knowledge Distillation}
Knowledge distillation \cite{gou2021knowledge} provides a strategy to transfer the knowledge learned by the teacher network to the student network, which has been applied in high-level computer vision tasks like object detection and image classification~\cite{wang2019distilling}. Recent work~\cite{hong2020dist} presents three challenges for applying knowledge distillation to the dehazing task. First, what kind of teacher task can help the dehazing task. Second, how the teacher network helps the dehazing network during training. Third, which similarity measure between teacher task and student task should be chosen. Fig. \ref{fig:kddn} shows the knowledge distillation strategy adopted by various dehazing algorithms. Different methods may use different numbers and locations of output features to compute feature loss.

KDDN~\cite{hong2020dist} designs a process-oriented learning mechanism, where the teacher network $T$ is an auto-encoder for high-quality haze-free image reconstruction. When training the dehazing network, the teacher network assists in feature learning, and optimizes $L_{T} = ||J(x) - T(J(x))||_{1}$. Therefore, the teacher task and the student task proposed by KDDN are two different tasks. In order to make full use of the feature information learned by the teacher network and help the training of the dehazing network. KDDN uses feature matching loss and haze density aware loss by a linear transformation function $g$, which are represented by (\ref{eq:l_rm}) and (\ref{eq:l_wrm}), respectively.
\begin{equation}
\label{eq:l_rm}
L_{rm} = \sum_{(m,n)\in{C}}|T^{m}(J(x)) - g(S^{n}(I(x)))|,
\end{equation}
\begin{equation}
\label{eq:l_wrm}
L_{wrm} = \sum_{(m,n)\in{C}}\psi \times {|T^{m}(J(x)) - g(S^{n}(I(x)))|},
\end{equation}
where $T^{m}$ represents the $m$-th layer of the teacher network, and the corresponding $S^{n}$ represents the $n$-th layer of the student network; $\psi$ is obtained by the normalization operation. KDDN can be trained without real transmission map, replaced by the residual between hazy and haze-free images.

KTDN~\cite{wu2020knowledge} is jointly trained using a teacher network and a dehazing network with the same structure. Through feature level loss, the prior knowledge possessed by the teacher network can be transferred to the dehazing network. SRKTDN~\cite{chen2021srktdn} uses ResNet18 pre-trained on ImageNet (with classification layers removed) as the teacher network, transferring many statistical experiences to the Res2Net101 encoder for dehazing. DALF~\cite{fang2021guiding} integrates dual adversarial training into the training process of knowledge distillation to improve the imitation ability of the student network to the teacher network. Applying knowledge distillation to dehazing networks provides a new and efficient way to introduce external prior knowledge.

\subsection{Transformation of Colorspace}
The input data of the dehazing network are usually three channel color images in RGB mode. By calculating the mean square error of hazy and haze-free images in RGB space and YCrCb space on the Dense-Haze dataset, Bianco et al.~\cite{bianco2019high} find that haze shows obvious numerical differences in the two spaces. The error values for the red, green and blue channels in RGB space are very close. However, in the YCrCb space, the error value of the luminance channel is significantly larger than that of the blue and red chroma components. AIP-Net~\cite{wang2018aipnet} performs a similar error comparison of color space transformation on synthetic datasets and obtained the same conclusion as ~\cite{bianco2019high}. Quantitative results~\cite{wang2018aipnet, bianco2019high} obtained by training the dehazing network in the YCrCb color space show that RGB space is not the only effective color mode for deep learning based dehazing methods. Furthermore, TheiaNet~\cite{mehra2021theianet} comprehensively analyzes the performance obtained by training the dehazing model in RGB, YCrCb, HSV and LAB color spaces. Experiments~\cite{singh2020single_BPP,sheng2022multi,chen2021srktdn,dudhane2019ryf} show that converting images from RGB space to other color spaces for model training is an effective scheme.

\subsection{Contrastive Learning}
In the process of training a non-ASM-based supervised dehazing network, a common way is to use the hazy image as the input of the network and expect to obtain a clear image. In this process, the clear image is used as a positive example to guide the optimization of the network. By designing pairs of positive and negative examples, AECR-Net provides a new perspective for treating hazy and haze-free images. Specifically, the clear image $J(x)$ and the dehazed image $J_{pred}(x)$ are taken as a positive sample pair, while the hazy image $I(x)$ and the dehazed image $J_{pred}{(x)}$ are taken as a negative sample pair. According to contrastive learning \cite{khosla2020supervised}, $J_{pred}(x)$, $J(x)$ and $I(x)$ can be regarded as anchor, positive and negative, respectively. For the pre-trained model $G(\cdot)$, the dehazing loss can be regarded as the sum of the reconstruction loss and the regularization term, as following:
\begin{equation}
\min{||J(x) - \psi(I(x))||_1} + \lambda \sum_{i=1}^{N}{\omega_i} \cdot \frac{||G_{i}(J(x)), G_{i}(\phi(I(x)))||_1}{||G_{i}(I(x)), G_{i}(\phi(I(x)))||_1},
\end{equation}
where $G_i$ represents the output feature of the $i$-th layer of the pre-trained model; $\lambda$ is the weight ratio between the image reconstruction loss and the regularization term; $w_i$ is weight factor for feature output; $\phi$ is the dehazing network. AECR-Net provides a universal contrastive regularization strategy for existing methods without adding extra parameters.

\subsection{Non-deterministic Output}
Dehazing methods based on deep learning usually set the optimization objective to obtain a single dehazed image. By introducing 2-dimensional latent tensors, pWAE~\cite{kim2021pixel} can generate different styles of dehazed images, which extends the general training purpose. pWAE proposes dehazing latent space $z_h$ and style latent space $z_s$ for dehazing, and applies the mean function $\mu(\cdot)$ and standard deviation function $\sigma(\cdot)$ to perform the transformation of the space:
\begin{equation}
z_{h\to{s}} = \sigma(z_s)(\frac{z_h - \mu{(z_h)}}{\sigma(z_h)}) + \mu{(z_s)}.
\end{equation}

A natural question is how to adjust the magnitude of the spatial mapping to control the degree of style transformation. pWAE uses linear module $z_{h}^{s} = \alpha{z_{h\to{s}}} + (1 - \alpha)z_h$ to adjust the weight between the dehazed image and the style information. By controlling $\alpha$, different degrees of stylized dehazed images corresponding to the style image can be obtained.

DehazeFlow~\cite{li2021dehazeflow} proposes that the dehazing task itself is ill-posedness, so it is unreliable to learn a model with deterministic one-to-one mapping. With a conditional normalizing flow network, DehazeFlow can compute the conditional distribution of clear images from a given hazy image. Therefore, it can learn a one-to-many mapping and obtain different dehazing results in the inference stage. The non-deterministic output~\cite{kim2021pixel,li2021dehazeflow} brings interpretable flexibility to the dehazing algorithm. Since the visual evaluation criteria of individual human beings are inherently different, one-to-many dehazing mapping provides more options for photographic work.

\subsection{Retinex Model}
Non-deep learning based research~\cite{galdran2018duality} demonstrates the duality between Retinex and image dehazing. Apart from the already widely used ASM, recent work~\cite{Li2021deep_retinex} explores the combination of Retinex model and CNN for dehazing. Retinex theory proposes that the image can be regarded as the element-wise product of the reflectance image $R$ and the illumination map $L$. Assuming the reflectance $R$ is illumination invariant, the relationship between hazy and haze-free image can be modeled by a retinex-based decomposition model as following:
\begin{equation}
\label{eq:retinex}
I(x) = J(x) \ast \frac{L_I}{L_J} = J(x) \ast L_{r},
\end{equation}
where $*$ means the multiplication in an element-wise way. $L_r$ can be seen as absorbed and scattered light caused by haze, which is determined by the ratio of the hazy image illumination map $L_{I}$ and natural illumination map $L_J$. 

Compared with ASM, which has been proven to be reliable, Retinex has a more compact physical form and fewer parameters to be estimated, but is not widely combined with CNN for dehazing, yet.
\subsection{Residual Learning}
Rather than directly learning the mapping from hazy to haze-free image, several methods argue that the residual learning method can reduce the learning difficulty of the network. The dehazing methods based on residual learning is performed at the image level. GCANet~\cite{chen2019Gated} uses the residual \{$J(x) - I(x)$\} between haze-free and hazy images as the optimization objective. DRL~\cite{du2018recursive}, SID-HL~\cite{xiao2020single} and POGAN~\cite{du2019recursive} believe that the way of residual learning is related to ASM, and the relationship can be obtained by reformulated ASM:
\begin{equation}
\begin{aligned}
I(x) &= J(x) + (A - J(x))(1 - t(x)) = J(x) + r(x),
\end{aligned}
\end{equation}
where $r(x)=(A - J(x))(1 - t(x))$ can be interpreted as a structural error term. By predicting nonlinear signal-dependent degradation $r(x)$, a clear image $J(x) = I(x) - r(x)$ can be obtained.

\subsection{Frequency Domain}
The dehazing methods based on CNN usually use down-sampling and upsampling for feature extraction and clear image reconstruction, and this process less considers the frequency information contained in the image. There are currently two approaches to combine frequency analysis and dehazing networks. The first is to embed the frequency decomposition function into the convolutional network, and the second is to use frequency decomposition as a constraint for loss computation.
For a given image, one low-frequency component and three high-frequency components can be obtained by wavelet decomposition. Wavelet U-net ~\cite{yang2019wavelet} uses discrete wavelet transform (DWT) and inverse discrete wavelet transform (IDWT) to facilitate high quality restoration of edge information in a clear image. Through the 1D scaling function $\phi(\cdot)$ and the wavelet function $\psi(\cdot)$, the 2D wavelets low-low (LL), low-high (LH), high-low (HL) and high-high (HH) are calculated as follows
\begin{gather}
\Phi_{LL}(m, n) = \phi(m)\phi(n), \notag \\
\Psi_{LH}(m, n) = \phi(m)\psi(n), \notag \\
\Psi_{HL}(m, n) = \psi(m)\phi(n), \notag \\
\Psi_{HH}(m, n) = \psi(m)\psi(n).
\end{gather}
where $m$ and $n$ represent the horizontal and vertical coordinates, respectively. MsGWN~\cite{dong2020deep} uses Gabor wavelet decomposition for feature extraction and reconstruction. By setting different degree values, the feature extraction module of MsGWN can obtain frequency information in different directions. EMRA-Net~\cite{wang2021ensemble} uses Haar wavelet decomposition as a downsampling operation instead of nearest downsampling and strided-convolution to avoid the loss of image texture details. By embedding wavelet analysis into convolutional network or loss function, recent studies~\cite{yang2020net,dharejo2021deep,fu2021dwgan} have shown that 2D wavelet transform can improve the recovery of high-frequency information in the wavelet domain. These works successfully apply wavelet theory and neural networks to dehazing tasks.

Besides, TDN~\cite{liu2020trident} introduces the Fast Fourier transform (FFT) loss as a constraint in the frequency domain. With the help of supervised training on amplitude and phase, the visual perceptual quality of images is improved without any additional computation in the inference stage.

\subsection{Joint Dehazing and Depth Estimation}
The scattering particles in the hazy environment will affect the accuracy of the depth information collected by the LiDAR equipment. S2DNet~\cite{hambarde2020s2dnet} proves that high-quality depth estimation algorithms can help the dehazing task. According to ASM, it can be known that the transmission map $t(x)$ and the depth map $d(x)$ have a negative exponential relationship $t(x)=e^{-\beta{d(x)}}$. Based on this physical dependency, SDDE~\cite{lee2020cnn} uses four decoders to integrate estimates of atmospheric light, clear image, transmission map, and depth map into an end-to-end training pipeline. In particular, SDDE proposes a depth-transmission consistency loss based on the observation that the standard deviation value ($std$) of the transmission map and the depth map pair should tend to zero:
\begin{equation}
L = ||std(\ln(t_{pred}{(x)} / d_{pred}{(x)}))||_2,
\end{equation}
where $t_{pred}{(x)}$ and $d_{pred}{(x)}$ represent the predicted transmission map and depth map, respectively. Aiming at better dehazing, TSDCN-Net ~\cite{cheng2021two} designs a cascaded network for two-stage methods with depth information prediction. Quantitative experimental results~\cite{yang2021depth} show that this joint estimation method can improve the accuracy of dehazing task.

\subsection{Segmentation and Detection with Dehazing}
Existing experiments~\cite{li2017all} show that the existence of image haze may cause various problems in detection algorithms, such as missing targets, inaccurate localizations and unconfident category prediction. Recent work~\cite{sakaridis2018model} has shown that haze also brings difficulties to semantic scene understanding. As a preprocessing module for high-level computer vision tasks, the dehazing process is usually separated from high-level computer vision tasks.

Li et al.~\cite{li2017all} jointly optimize the object detection algorithm and AOD-Net, and the result proves that the dehazing algorithm can promote the detection task. LEAAL~\cite{li2020deep} proposes that fine-tuning the parameters of the object detector during joint training may lead to a detector biased towards the haze-free images generated by the pretrained dehazing network. Different from the fine-tuning operation of Li et al.~\cite{li2017all}, LEAAL uses object detection as an auxiliary task for dehazing and the parameters of the object detector are not fully updated during the training process.

UDnD~\cite{zhang2020unified} takes advantage of each other by jointly training a dehazing network and a dense-aware multi-domain object detector. The object detection network is trained by adopting the classification and localization terms used by Region Proposal Network and Region of Interest. The multi-task training approach used by UDnD can consider the reduced inter-domain gaps and the remained intra-domain gaps for different density levels.

Recent work has explored performing dehazing and high-level computer vision tasks simultaneously without using ASM. SDNet~\cite{zhang2021semantic} combines semantic segmentation and dehazing into a unified framework in order to use semantic prior as a constraint for the optimization process. By embedding the predictions of the segmentation map into the dehazing network, SDNet performs a joint optimization of pixel-wise classification loss and regression loss. The classification loss is
\begin{equation}
L_{sem}(s, s^{*}) = - \frac{1}{P}\sum_{i}s_{i}^{*}{\log(s_{i})},
\end{equation}
where $P$ is the total number of pixels; $s_i$ is the class prediction at position $i$; $s^*$ denotes the ground truth semantic annotation.

~\cite{zhang2021semantic,zhang2020unified,li2020deep} show that segmentation and detection can be performed by embedding with dehazing networks. The way of joint dehazing task and high-level vision task may reduce the computational load to a certain extent by sharing the learned features, which can expand the goal of dehazing research.

\subsection{End-to-end CNN}
The ``end-to-end'' CNN stands for the non-ASM-based supervised algorithms, which usually consist of well-designed neural networks that take a single hazy image as input and a haze-free image as output. Networks based on different ideas are adopted, which are summarized as follows.
\begin{itemize}
\item Attention mechanism: FFA-Net~\cite{qin2020FFA}, GridDehazeNet~\cite{liu2019grid}, SAN~\cite{liang2019selective}, HFF~\cite{zhang2021hierarchical_tcsvt}.
\item Encoder-Decoder: CAE~\cite{chen2019convolutional}.
\item Based on dense block: 123-CEDH~\cite{guo2019dense123}.
\item U-shaped structure: DSEU~\cite{lee2020image}, MSBDN~\cite{dong2020multi}.
\item Hierarchical network: DMHN~\cite{das2020fast}.
\item Fusion with bilateral grid learning: 4kDehazing~\cite{zheng2021ultra}.
\end{itemize}

The end-to-end dehazing networks have an important impact on the entire dehazing field, proving that numerous deep learning models are beneficial to the dehazing task.

\section{Semi-supervised Dehazing}
\label{sec:semi_supervised}
Compared with the research on supervised methods, semi-supervised dehazing~\cite{zhao2021refinednet,chen2021psd} algorithms have received relatively less attention. An important advantage of semi-supervised methods is the ability to utilize both labeled and unlabeled datasets. Therefore, compared to the fully supervised dehazing model, the semi-supervised dehazing model can alleviate the requirement of paired hazy and haze-free images. For dehazing task, labeled datasets are usually synthetic or artificially generated, while unlabeled datasets usually contain real world hazy images. According to the analysis of the dataset in Section \ref{sec:related_work}, there are inherent differences between synthetic data and real world data. Therefore, semi-supervised algorithms usually have the ability to mitigate the gaps between synthetic domain and the real domain.

\begin{figure}
\centering
\includegraphics[width=14cm,height=3.24cm]{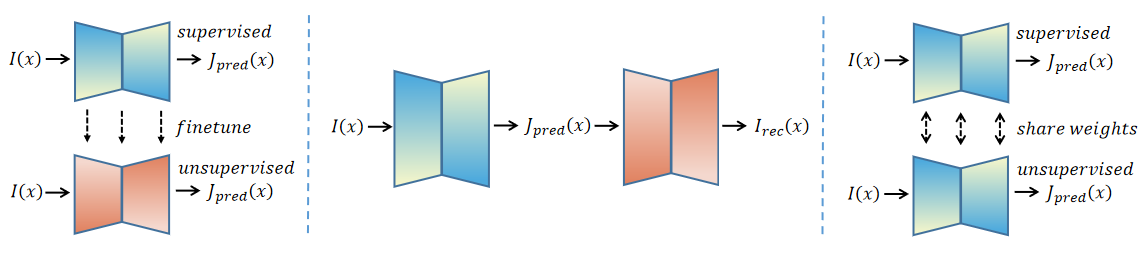}
\\
(a) Pretrain and finetune \hspace{1.3cm} (b) Disentangled and reconstruction \hspace{1cm} (c) Two branches training
\caption{Semi-supervised dehazing methods.}
\label{fig:semi_supervised}
\end{figure}

Fig. \ref{fig:semi_supervised} shows the principles of various semi-supervised dehazing models, where the network is a schematic diagram.

\subsection{Pretrain Backbone and Finetune}
PSD~\cite{chen2021psd} proposes a domain adaptation method that can be extended by existing models. It first uses a powerful backbone network for pre-training to obtain a base model suitable for synthetic data. Then the fine-tuning of the real domain in an unsupervised manner is applied to the well-trained network, thereby improving the generalization ability of the model to real world hazy images. To achieve fine-tuning in the real domain, PSD combines DCP loss, Bright Channel Prior (BCP)~\cite{wang2013automatic} loss and Contrast Limited Adaptive Histogram Equalization (CLAHE) loss together. BCP loss and CLAHE loss are
\begin{equation}
L_{BCP} = ||t_{BCP}(x) - t_{pred}(x)||_{1},
\end{equation}
\begin{equation}
L_{CLAHE} = ||I(x) - I_{CLAHE}(x)||_{1},
\end{equation}
where $t_{BCP}$ represents the transmission map estimated by BCP and $t_{pred}$ is the predicted output of the network. $I_{CLAHE}$ is the hazy image reconstructed by CLAHE and other physical parameters.

During fine-tuning, the model may forget the useful knowledge it learned during the pre-training phase. Therefore, PSD proposes a feature-level constraint, which is achieved by calculating the feature map difference between the fine-tuning network and the pre-trained network.  By feeding the synthetic data and real data into the fine-tuning network and the pre-trained network, respectively, four feature maps can be obtained: $F_{syn}^{tune}$, $F_{syn}^{pre}$, $F_{real}^{tune}$, and $F_{real}^{pre}$. Then, the loss for preventing knowledge forgetting can be calculated as
\begin{equation}
L_{lwf} = ||F_{syn}^{tune} - F_{syn}^{pre}||_{1} + ||F_{real}^{tune} - F_{real}^{pre}||_{1}.
\end{equation}
As shown in Fig. \ref{fig:semi_supervised}(a), supervised pre-training is performed first, and then the dehazing network is fine-tuned in an unsupervised form.

SSDT~\cite{zhang2021single_ssdt} uses the encoder-decoder network for pre-training in the form of domain translation. After pre-training, two encoders and one decoder are selected to obtain the holistic dehazing network $G(\cdot)$. Then, $G(\cdot)$ is fine-tuned with synthetic hazy images and real world hazy images. The two-stage method described above needs to ensure that the pre-training process can meet the accuracy requirements, otherwise it may bring accumulated errors to the fine-tuning process.

\subsection{Disentangled and Reconstruction}
From the perspective of dual learning, the dehazing task and the haze generation task can be able to assist each other. Based on this assumption, DCNet~\cite{chen2021dcnet} proposes a dual-task cycle network that jointly utilizes labeled dataset $N$ and unlabeled dataset $M$ by dehazing network $DN(\cdot)$ and haze generation network $HGN(\cdot)$. The total loss is combine by dehazing loss $L_{DN}$ and reconstruction loss $L_{HGN}$, that is $L_{total} = L_{DN} + L_{HGN}$, as
\begin{equation}
\begin{aligned}
L_{total} = &\sum_{i=1}^{M+N}P(I_i(x))L(DN(I_i(x)), J_i(x), \epsilon_{1}) + \lambda{L(HGN(DN(I_i(x))), I_i(x), \epsilon_{2})},
\end{aligned}
\end{equation}
where $\epsilon_1$ and $\epsilon_2$ are regularizing hyper-parameter; $P(I_i(x))$ equals $1$ when the $I_i(x)$ comes from the labeled dataset $N$, and equals $0$ otherwise. As shown in Fig. \ref{fig:semi_supervised} (b), the predicted haze-free image $J_{pred}(x)$ is first obtained by the left network, and then the hazy image $I_{rec}(x)$ is reconstructed by the right network.

Liu et al.~\cite{liu2021synthetic} use a disentangled image dehazing network (DID-Net) and a disentangled-consistency mean-teacher network (DMT-Net) to combine labeled and unlabeled data. DID-Net is responsible for disentangling the hazy image into a haze-free image, the transmission map, and the global atmospheric light. DMT-Net is used to jointly exploit the labeled synthetic data and unlabeled real world data through disentangled consistency loss. The supervised loss consists of four terms: haze-free image prediction $L_J^s$, transmission map prediction $L_t^s$, atmospheric light prediction $L_A^s$, and hazy image reconstruction $L_{rec}$:
\begin{gather}
L_J^s = ||G_J - P_J||_1 + ||G_J - \hat{P}_J||_1, \notag \\
L_T^s = ||G_T - P_T||_1 + ||G_T - \hat{P}_T||_1, \notag \\
L_A^s = ||G_A - P_A||_1 + ||G_A - \hat{P}_A||_1, \notag \\
L_{rec} = ||I(x) - P_{I}||_1 + ||I(x) - \hat{P}_I||_1,
\end{gather}
where $G$ stands for ground-truth, $P$ is the prediction of the first stage, and $\hat{P}$ is the prediction of the second stage. The supervised loss $L^{s}(x)$ is the weighted sum of the above four losses, that is $L^{s}(x) = L_{J}^s + \alpha_1 L_T^s + \alpha_2 L_A^s + \alpha_3 L_{rec}$. For unlabeled data, consistency loss is used to constrain the teacher network $T$ and the student network $S$:
\begin{gather}
L_{J}^{c} = ||S_J - T_J||_1 + ||S_{\hat{J}} - T_{\hat{J}}||_1, \notag \\
L_{T}^{c} = ||S_T - T_T||_1 + ||S_{\hat{T}} - T_{\hat{T}}||_1, \notag \\
L_{A}^{c} = ||S_A - T_A||_1 + ||S_{\hat{A}} - T_{\hat{A}}||_1, \notag \\
L_{rec}^{c} = ||S_I - T_I||_1 + ||S_{\hat{I}} - T_{\hat{I}}||_1,
\end{gather}
where the subscript $\hat{J}$, $\hat{T}$, $\hat{A}$ and $\hat{I}$ denote the results predicted in second stage. Thus, the loss for the unlabeled dataset is $L^{c}(y) = L_J^c + \alpha_{4} L_T^c + \alpha_{5} L_A^c + \alpha_{6}L_{rec}^c$. The final loss function of the semi-supervised framework consists of a supervised loss on the labeled dataset $N$ and a consistency loss on the unlabeled dataset $M$, that is $L_{total} = \sum_{x \in N}{L^s{(x)}} + \lambda \sum_{y \in M}{L^c{(y)}}$.

CCDM~\cite{zhang2020color} designs a color-constrained dehazing model that can be extended to a semi-supervised framework, which is achieved by the reconstruction of hazy images, the smoothing loss of $t(x)$ and $A$, etc. Experiments~\cite{liu2021synthetic,zhang2020color,chen2021dcnet} show that the reconstruction of hazy images can provide effective supervisory signals in an unsupervised manner, which is instructive for semi-supervised frameworks.

\subsection{Two-branches Training}
SSID~\cite{Li2019semi} designs an end-to-end network that integrates a supervised learning branch and an unsupervised learning branch. The training process of SSID uses both the labeled dataset and the unlabeled dataset by the following process:
\begin{equation}
J_{pred}(x) = G(I(x)),
\end{equation}
where $G(\cdot)$ consists of a supervised part $G_{s}(\cdot)$ and an unsupervised part $G_{u}(\cdot)$. The supervised loss composed of L2 loss and perceptual loss is used to ensure that the predicted image $J_{pred}(x)$ and its corresponding ground truth image are as close as possible, which is the same as the supervised dehazing algorithm. A combination of total variation loss and dark channel loss is used for unsupervised training. As shown in Fig. \ref{fig:semi_supervised}(c), supervised training and unsupervised training are performed in a shared weight manner.

SSIDN~\cite{an2021semi} also combines supervised and unsupervised training processes. Supervised training is used to learn the mapping from hazy to haze-free images. With dark channel prior and bright channel prior~\cite{wang2013automatic} guiding the training process, the unsupervised branch incorporates the estimation of the transmission map and atmospheric light.

The DAID~\cite{shao2020domain} adopts a domain adaptation model to jointly train multi-subnetworks. The $G_{S\to{R}}$ and $G_{R\to{S}}$ are the image translation modules used for the translation between the synthetic domain and the real domain, where $R$ and $S$ stand for real domain and synthetic domain, respectively. By using an image-level discriminator $D_{R}^{img}$ and a feature-level discriminator $D_{R}^{feat}$, the adversarial loss of the translation process can be calculated. In order to ensure that the content of images are maintained during the translation process, DAID uses cycle consistency loss to constrain the translation network. Furthermore, identity mapping loss is also used in the conversion process to restrict the identity of the image generation process in two domains. The training of the dehazing network $G_{R}$ is a combination of unsupervised and supervised processes. The supervised process minimizes the $L_{rm}$ loss to make the dehazed image $J_{S\to{R}}$ closer to the corresponding haze-free image $Y_{S}$:
\begin{equation}
L_{rm} = ||J_{S\to{R}} - Y_{S}||_{2}^{2}.
\end{equation}

Then, the total variation loss and dark channel loss are used as unsupervised losses. For the training of the dehazing network $G_S$, there is also a combination of supervised loss and unsupervised loss. With the help of domain transformation and unsupervised loss, DAID can effectively reduce the gap between the synthetic domain and the real domain.

As introduced above~\cite{Li2019semi,an2021semi,shao2020domain}, using supervised branch and unsupervised branch for joint training to build a semi-supervised framework can effectively alleviate the problem of domain shift.

\section{Unsupervised Dehazing}
\label{sec:unsupervised}
The supervised and semi-supervised dehazing methods have achieved excellent performance on public datasets. However, the training process requires paired data (i.e. hazy images and haze-free images / transmission maps), which are difficult to obtain in the real world. For outdoor scenes containing grass, water or moving objects, it is difficult to guarantee that two images taken under hazy and clear weather have exactly the same content. If the haze-free labels are not accurate enough, the accuracy of dehazed images will be reduced. Therefore, ~\cite{engin2018cycle,dudhane2019cdnet,liu2020end,wei2021sidgan} explore dehazing algorithms in an unsupervised way.

\begin{figure}
\centering
\includegraphics[width=15cm,height=2.5cm]{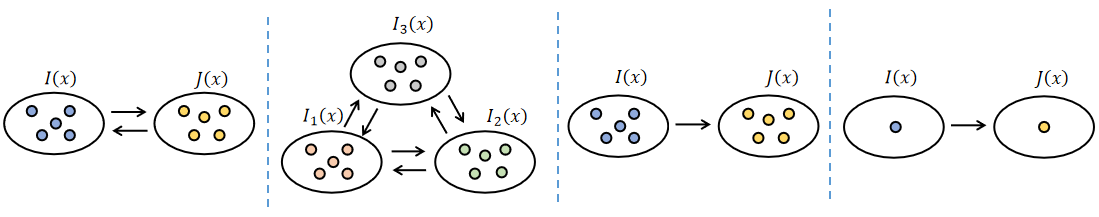}
\\
(a) Between two domains \hspace{0.5cm}(b) Among multi-domains \hspace{1.3cm}(c) Without target \hspace{1.5cm} (d) Zero-shot \hspace{0.7cm}
\caption{Schematic diagram of the unsupervised dehazing algorithms.}
  \label{fig:unsupervised}
\end{figure}

\subsection{Unsupervised Domain Transfer}
In the study of image style transfer and image-to-image translation, CycleGAN~\cite{zhu2017unpaired} is proposed, which provides a way for learning the bidirectional  mapping functions between two domains. Inspired by CycleGAN, \cite{engin2018cycle,dudhane2019cdnet,liu2020end} are designed for unsupervised transformation of hazy and haze-free / transmission domains. The Cycle-Dehaze~\cite{engin2018cycle} contains two generators $G$ and $F$, which are used to learn the mapping from hazy domain to haze-free domain and the reverse mapping, respectively. As shown in Fig. \ref{fig:unsupervised} (a), the hazy and haze-free images are translated to each other by two generators. By sampling $x$ and $y$ from the hazy domain $X$ and the haze-free domain $Y$, the Cycle-Dehaze uses the perceptual metric (denoted as $\psi(\cdot)$) to obtain the cyclic perceptual consistency loss:
\begin{equation}
\begin{aligned}
L_{cyc-p} = &{||\psi(x) - \psi(F(G(x)))}{||}_{2}^2 + {||\psi(y) - \psi(G(F(y)))}{||}_{2}^2.
\end{aligned}
\end{equation}

The overall loss function of Cycle-Dehaze is composed of cyclic perceptual consistency loss and Cycle-GAN's loss function, which can alleviate the requirement of paired data. CDNet~\cite{dudhane2019cdnet} also adopts a cycle-consistent adversarial approach for unsupervised dehazing network training. Unlike Cycle-Dehaze, CDNet embeds ASM into the network architecture for estimating transmission map, which enables it to acquire physical parameters while restoring haze-free images. E-CycleGAN~\cite{liu2020end} adds ASM and a priori statistical law to estimate atmospheric light on the basis of CycleGAN, which allows it to perform independent parameter estimation for sky regions.

USID~\cite{huang2019towards} introduces the concept of haze mask $H(x)$ and constant bias $\epsilon$, which can obtain the reformulated ASM as
\begin{equation}
J(x) = \frac{I(x) - A}{H(x)} + A + \epsilon.
\end{equation}
By combining with cycle consistent loss, USID also removes the requirement of an explicit paired haze/depth data in an unsupervised multi-task learning manner. In order to decouple content and haze information, USID-DR~\cite{liu2019unsupervised} designs a content encoder and a haze encoder embedded in CycleGAN. This decoupling approach can enhance feature extraction and reconstruction during cycle consistent training. USID-DR proposes that making the output of the encoder conform to the Gaussian distribution by latent regression loss can improve the quality of hazy image synthesis process, thereby improving the overall dehazing performance. DCA-CycleGAN~\cite{mo2021dca} utilizes the dark channel to build the input and generates the attention for handling the nonhomogeneous haze. These CycleGAN-based methods demonstrate that unsupervised hazy and haze-free image domain transformation can achieve the same performance as supervised algorithms from both ASM-based and non-ASM-based perspectives.

For the dehazing task, the haze density in hazy images can be very different. Most supervised, semi-supervised and unsupervised dehazing methods hold the view that hazy images and haze-free images should be treated as two domains, without considering the density among different examples. Recent work~\cite{jin2020unsupervised} proposes that it is an important issue to apply the haze density information to the unsupervised training process, thereby improving the generalization ability of the dehazing model to images with different densities of haze. An unsupervised conditional disentangle network, called UCDN~\cite{jin2020unsupervised}, is designed by incorporating conditional information into the training process of CycleGAN. Further, DHL-Dehaze~\cite{cong2020discrete} analyzes multiple haze density levels of hazy images, and proposes that the difference in haze density should be fully utilized in the dehazing procedure. Compared to UCDN which contains four submodules, DHL-Dehaze has only two networks to train. The idea of DHL-Dehaze is based on the research of multi-domain image-to-image translation. As shown in Fig. \ref{fig:unsupervised}(b), the images $I_1(x)$, $I_2(x)$ and $I_3(x)$ of different densities can be transformed to other domains, where $I_{3}(x)$ can be explained as the haze-free domain. The training of DHL-Dehaze consists of a adversarial process and a classification process. By sampling $X_{ori}=\{x_{ori}^{(1)}, x_{ori}^{(2)}, \ldots, x_{ori}^{(m)}\}$ with corresponding density labels $L_{ori} = \{l_{ori}^{(1)}, l_{ori}^{(2)}, \ldots, l_{ori}^{(m)}\}$, the classification loss $L_{cls}^{ori}$ and adversarial loss $L_{dis}^{ori}$ in the source domain can be obtained by (\ref{eq:L_cls_ori}) and (\ref{eq:L_dis_ori}):
\begin{equation}
\label{eq:L_cls_ori}
L_{cls}^{ori}=\frac{1}{m}\sum_{i=1}^{m}-logD_{cls}(l_{ori}^{(i)}\mid x_{ori}^{(i)}),
\end{equation}
\begin{equation}
\label{eq:L_dis_ori}
L_{dis}^{ori}=\frac{1}{m}\sum_{i=1}^{m}-logD_{dis}(x_{ori}^{(i)}).
\end{equation}

In order to generate the images of the target domain, the labels of target domain $L_{tar} = \{l_{tar}^{(1)}, l_{tar}^{(2)}, \ldots, l_{tar}^{(m)}\}$ are required. DHL-Dehaze sends $X_{ori}$ and $L_{tar}$ into the multi-scale generator (MST) together, and obtains new images with the same attributes as $L_{tar}$:
\begin{equation}
\label{eq:x_tar}
X_{tar} = MST(X_{ori}, L_{tar}).
\end{equation}
Then, the classification loss $L_{cls}^{tar}$ and adversarial loss $L_{dis}^{tar}$ of the target domain image can be obtained by using (\ref{eq:L_cls_tar}) and (\ref{eq:L_dis_tar}):
\begin{equation}
\label{eq:L_cls_tar}
L_{cls}^{tar}=\frac{1}{m}\sum_{i=1}^{m}-logD_{cls}(l_{tar}^{(i)}\mid x_{tar}^{(i)}),
\end{equation}
\begin{equation}
\label{eq:L_dis_tar}
L_{dis}^{tar} = \frac{1}{m}\sum_{i=1}^{m}logD_{dis}(x_{tar}^{(i)}).
\end{equation}

It is worth noting that UCDN and DHL-Dehaze do not use the hazy and haze-free images of the same scene for supervision, but apply the density of haze as the supervisory information for the training process.

The training of unsupervised domain transformation algorithms is more difficult than that of supervised algorithms. The convergence of GAN-based domain transformation algorithms is difficult to determine, which may lead to over-enhanced images.

\subsection{Learning without Haze-free Images}
The training process of CycleGAN-based methods and DHL-Dehaze does not require paired data, which has greatly reduced the difficulty of data collection. Further, Deep-DCP~\cite{alona2020un_dcp} proposes to use only hazy image for training process. The strategy of domain translation dehazing algorithms is to learn the mapping relationship between hazy and haze-free / transmission domains, while the main idea of Deep-DCP is to minimize the DCP~\cite{he2010single} energy function. According to statistical assumption, the transmission map can be estimated in an unsupervised way. Based on the estimated transmission map and soft matting, the energy function $E(t_{\theta},I(x))$ can be obtained, where $\theta$ is the parameters for tuning. Thus, the goal of training is to minimize the energy function:
\begin{equation}
\theta^{*} = \mathop{\arg\min}_{\theta}[\frac{1}{N}\sum_{i=1}^{N}E(t_{\theta}(I_{i}(x)),I_{i}(x))].
\end{equation}

As shown in Fig. \ref{fig:unsupervised}(c), the network can automatically learn the mapping from hazy to haze-free images without requiring target domain labels.

\subsection{Unsupervised Image Decomposition}
Double-DIP~\cite{gandelsman2019double} proposes an unsupervised hierarchical decoupling framework based on the observation that the internal statistics of a mixed layer is more complex than the single layer that composes it. Suppose $Z$ is a linear sum of independent random variables $X$ and $Y$. From a statistical point of view, the entropy of $Z$ is larger than its independent components, that is, $H(Z) \geq \max{\{H(X),H(Y)\}}$. Based on this, Double-DIP proposes the loss for image layer decomposition:
\begin{equation}
L = L_{reconst} + \alpha \cdot L_{excl} + \beta \cdot L_{reg},
\end{equation}
where $L_{reconst}$ represents the reconstruction loss of the hazy image, $L_{excl}$ is the exclusion loss between the two DIPs, and $L_{reg}$ is the regularization loss used to obtain the continuous and smooth transmission map.

\subsection{Zero-Shot Learning for Dehazing}
Data-driven unsupervised dehazing methods have achieved impressive performance. Unlike those models that require sufficient data to perform network training, ZID~\cite{Li2020Zero}  proposes a neural network dehazing process that only requires a single example. ZID has further reduced the dependence of the parameter learning process on data by combining the advantages of unsupervised learning and zero-shot learning. Three sub-networks $f_{J}(\cdot)$ (J-Net), $f_{T}(\cdot)$ (T-Net) and $f_{A}(\cdot)$ (A-Net) are used to estimate the $J(x)$, $t(x)$ and $A$, respectively. By the reconstruction process, the hazy image $I(x)$ can be disentangled by minimizing $L_{rec}$ loss:
\begin{equation}
L_{rec} = ||I_{rec}(x) - I(x)||_{p},
\end{equation}
where $I_{rec}(x)$ is reconstructed hazy image, and $p$ denotes $p$-norm. The disentangled atmospheric light $f_{A}(x)$ and Kullback-Leibler divergence are used to obtain the atmospheric light $A$ as shown in the following formula:
\begin{equation}
\label{ZID_L_A}
\begin{aligned}
L_A = & L_H + L_{KL} = ||f_{A}(x) - A(x)||_{p} + KL(N(u_z, \delta_z^2)||N(0,I)),
\end{aligned}
\end{equation}
where $L_{H}$ is the loss for $f_{A}(x)$ and initial hint value $A(x)$ is automatically learned from data. It should be noted that $A(x)$ in ZID is not the same as $A$ in ASM; $N(0, I)$ stands for Gaussian distribution; $z$ is learned from input $x$. The unsupervised channel loss $L_J$ used in J-Net is for the decomposing process of the haze-free image $J(x)$. The $L_{J}$ is calculated based on the dark channel, and the formula is $L_{J} = ||\min_{c \in \{r, g, b\}}(J^{c}(y))||_{p}$, where $c$ denotes color channel and $y$ stands for local patch of the J-Net output. The purpose of $L_{reg}$ is to enhance the stability of the model and make $A$ and $t(x)$ smooth. The overall loss function of ZID is
\begin{equation}
L = L_{rec} + L_A + L_J + L_{reg}.
\end{equation}

YOLY~\cite{li2021yoly} uses three joint disentanglement subnetworks for clear image and physical parameter estimation, enabling unsupervised and untrained haze removal. As shown in Fig. \ref{fig:unsupervised}(d), ZID and YOLY can learn the mapping from hazy to haze-free image using a single unlabeled example.

The limitation of the zero-shot algorithm is that the generalization ability is limited, and the network must be retrained for each unseen example to get good dehazing performance.
\section{Experiment and Performance Analysis}
\label{sec:experiment}
This section provides quantitative and qualitative analysis of the dehazing performance of the baseline supervised, semi-supervised and unsupervised algorithms. The training examples used in the experiments are the indoor data ITS and outdoor data OTS from RESIDE~\cite{Li2019bench}. In order to accurately compare the dehazed images and the real clear images, the indoor and outdoor images included in the SOTS provided by RESIDE are used as the test set. Aiming at ensuring the fairness and representativeness of the experimental results, 10 representative algorithms are selected for comparison:
\begin{itemize}
\item supervised: AOD-Net~\cite{li2017all}, 4kDehazing~\cite{zheng2021ultra}, DMMFD~\cite{deng2019deep}, FFA-Net~\cite{qin2020FFA},  GCA-Net~\cite{chen2019Gated}, GridDehazeNet~\cite{liu2019grid},  MSBDN~\cite{dong2020multi}.
\item semi-supervised: SSID~\cite{Li2019semi}.
\item unsupervised: Cycle-Dehaze~\cite{engin2018cycle},  ZID~\cite{Li2020Zero}.
\end{itemize}

The framework used for the algorithm implementation is PyTorch, and the running platform is NVIDIA Tesla V100 32 GB (2 GPUs). Except for the zero-shot ZID, all algorithms use a batch size of 8 in the training phase. For the ITS and OTS datasets, the training epochs are set according to~\cite{qin2020FFA} and~\cite{liu2019grid}, respectively. For the supervised dehazing methods, four loss function strategies are adopted, including L1, L2, L1 + P, L2 + P, where P denotes the perceptual loss. For semi-supervised and unsupervised methods, the loss functions follow the settings in the respective papers. In the experiment, we first compared the convergence curves of different supervised algorithms when they were set to 0.001, 0.0005 or 0.0002, and selected the value with the best convergence effect as the best learning rate. After the training process, the loss values of all supervised models have stably converged. For semi-supervised and unsupervised algorithms, the learning rates used are those recommended in the corresponding papers.

In order to ensure as few interference factors as possible in the experiments, the following strategies are not used in the experiments: (1) pre-training, (2) dynamic learning rate adjustment, (3) larger batch size, and (4) data augmentation. Therefore, the quantitative results obtained in the following experiments may be a little lower than the best results provided in the papers corresponding to the various algorithms.

Tables \ref{tab:results_RESIDE_indoor} and \ref{tab:results_RESIDE_outdoor} show the PSNR/SSIM values obtained on the indoor and outdoor test sets of various algorithms on RESIDE SOTS, where the highest values are shown in bold. It can be concluded that FFA-Net achieves the best performance on the indoor test set, with the highest PSNR and SSIM. The best performance among different algorithms on the outdoor test set is achieved by different methods. It can be also seen from Tables \ref{tab:results_RESIDE_indoor} and \ref{tab:results_RESIDE_outdoor} that the perceptual loss has a certain effect on the performance of the model, but it is not obvious.

\begin{figure}
\centering
\hspace{0.4pt}
\includegraphics[width=2.2cm,height=1.5cm]{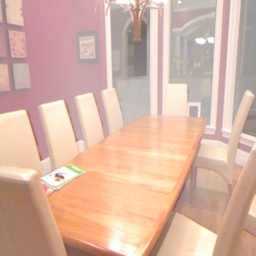}
\includegraphics[width=2.2cm,height=1.5cm]{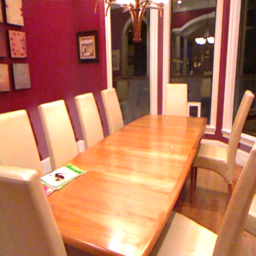}
\includegraphics[width=2.2cm,height=1.5cm]{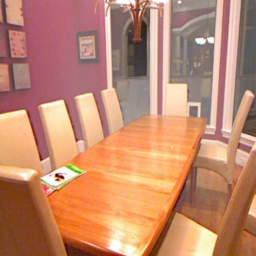}
\includegraphics[width=2.2cm,height=1.5cm]{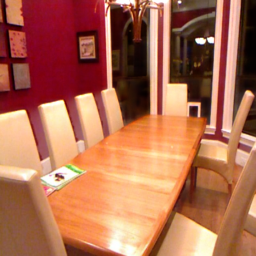}
\includegraphics[width=2.2cm,height=1.5cm]{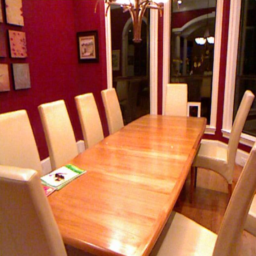}
\includegraphics[width=2.2cm,height=1.5cm]{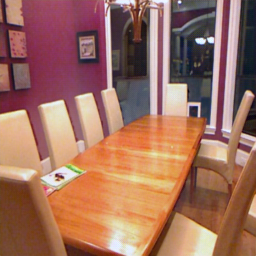}
\vspace{2pt}

\includegraphics[width=2.2cm,height=1.5cm]{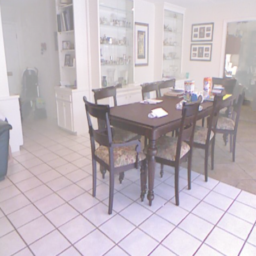}
\includegraphics[width=2.2cm,height=1.5cm]{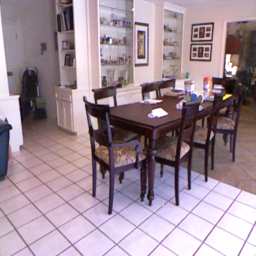}
\includegraphics[width=2.2cm,height=1.5cm]{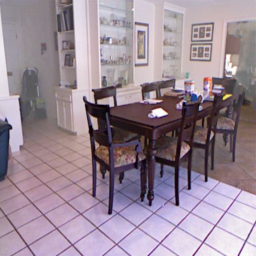}
\includegraphics[width=2.2cm,height=1.5cm]{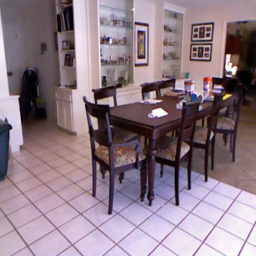}
\includegraphics[width=2.2cm,height=1.5cm]{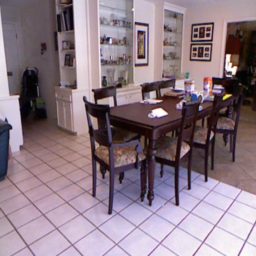}
\includegraphics[width=2.2cm,height=1.5cm]{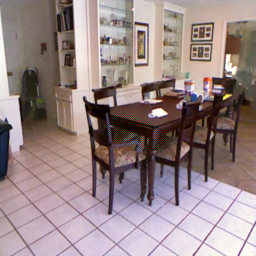}

\hspace{0.3cm} (a) hazy \hspace{0.5cm}  (b) 4kDehazing  \hspace{0.5cm}   (c) AODNet  \hspace{0.5cm}  (d) DMMFD  \hspace{0.8cm} (e) FFA-Net  \hspace{0.5cm}  (f) GCA-Net  \hspace{0.5cm}  \\
\vspace{2pt}
\hspace{0.4pt}
\includegraphics[width=2.2cm,height=1.5cm]{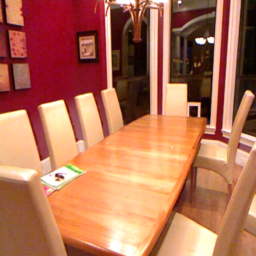}
\includegraphics[width=2.2cm,height=1.5cm]{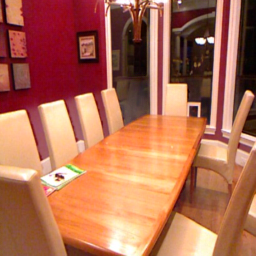}
\includegraphics[width=2.2cm,height=1.5cm]{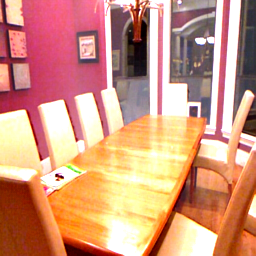}
\includegraphics[width=2.2cm,height=1.5cm]{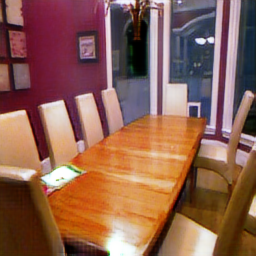}
\includegraphics[width=2.2cm,height=1.5cm]{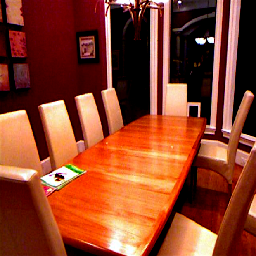}
\includegraphics[width=2.2cm,height=1.5cm]{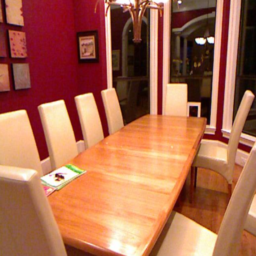}
\vspace{1.5pt}

\includegraphics[width=2.2cm,height=1.5cm]{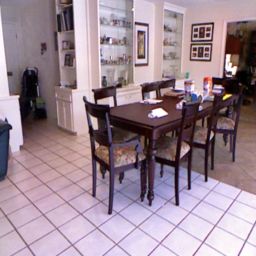}
\includegraphics[width=2.2cm,height=1.5cm]{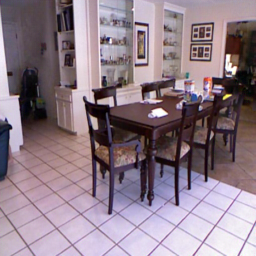}
\includegraphics[width=2.2cm,height=1.5cm]{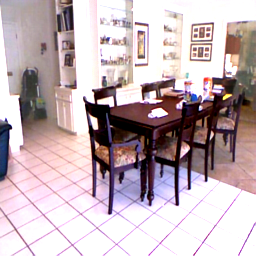}
\includegraphics[width=2.2cm,height=1.5cm]{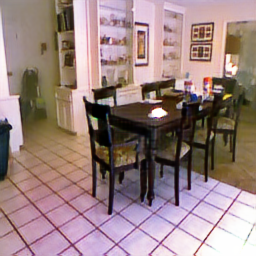}
\includegraphics[width=2.2cm,height=1.5cm]{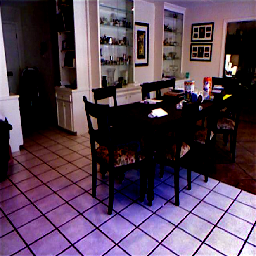}
\includegraphics[width=2.2cm,height=1.5cm]{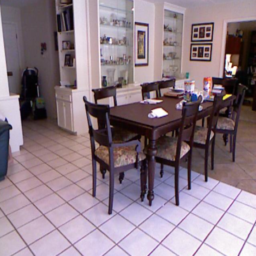}

\leftline{\hspace{1.0cm} (g) GridDehazeNet \hspace{0.1cm}  (h) MSBDN  \hspace{0.7cm}   (i) SSID  \hspace{0.7cm}  (j) Cycle-Dehaze  \hspace{0.7cm} (k) ZID  \hspace{0.9cm}  (l) clear  \hspace{0.5cm}}
\caption{Visual results on RESIDE SOTS indoor.}
 \label{fig:vis_indoor}
\end{figure}

\begin{figure}
\centering
\hspace{0.4pt}
\includegraphics[width=2.2cm,height=1.5cm]{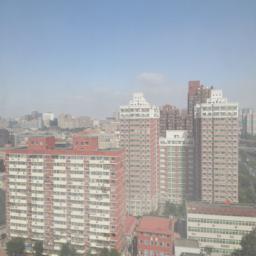}
\includegraphics[width=2.2cm,height=1.5cm]{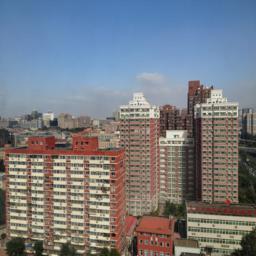}
\includegraphics[width=2.2cm,height=1.5cm]{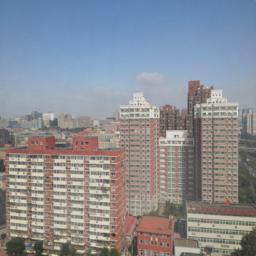}
\includegraphics[width=2.2cm,height=1.5cm]{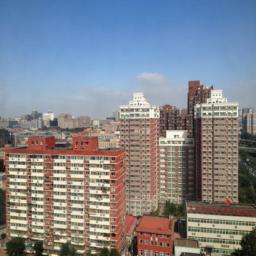}
\includegraphics[width=2.2cm,height=1.5cm]{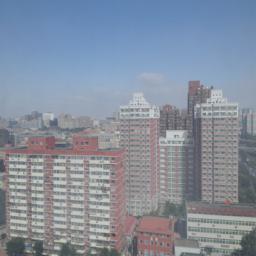}
\includegraphics[width=2.2cm,height=1.5cm]{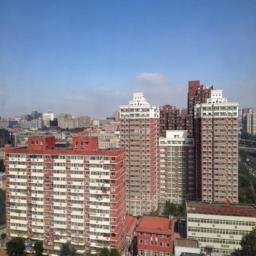}
\vspace{2pt}

\includegraphics[width=2.2cm,height=1.5cm]{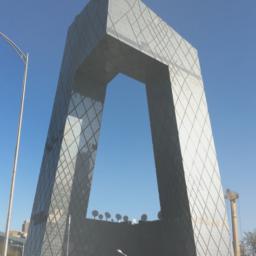}
\includegraphics[width=2.2cm,height=1.5cm]{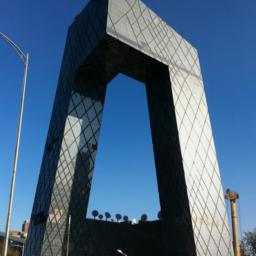}
\includegraphics[width=2.2cm,height=1.5cm]{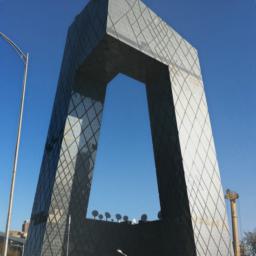}
\includegraphics[width=2.2cm,height=1.5cm]{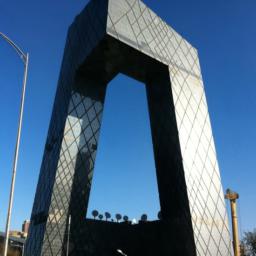}
\includegraphics[width=2.2cm,height=1.5cm]{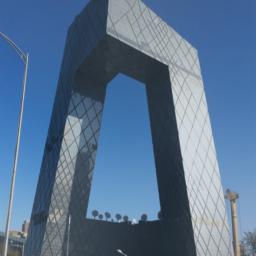}
\includegraphics[width=2.2cm,height=1.5cm]{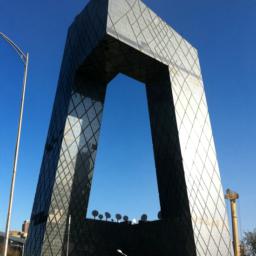}
\\
 \hspace{0.3cm} (a) hazy \hspace{0.5cm}  (b) 4kDehazing  \hspace{0.5cm}   (c) AODNet  \hspace{0.5cm}  (d) DMMFD  \hspace{0.8cm} (e) FFA-Net  \hspace{0.5cm}  (f) GCA-Net  \hspace{0.5cm}  \\
\vspace{2pt}
\includegraphics[width=2.2cm,height=1.5cm]{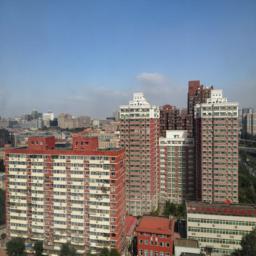}
\includegraphics[width=2.2cm,height=1.5cm]{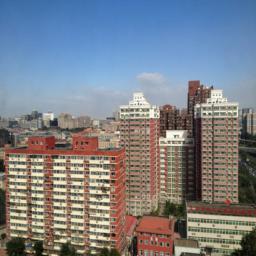}
\includegraphics[width=2.2cm,height=1.5cm]{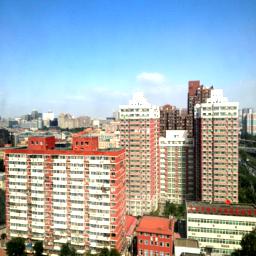}
\includegraphics[width=2.2cm,height=1.5cm]{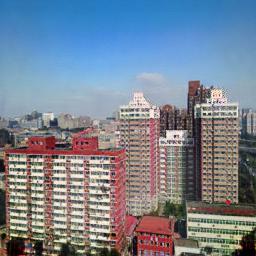}
\includegraphics[width=2.2cm,height=1.5cm]{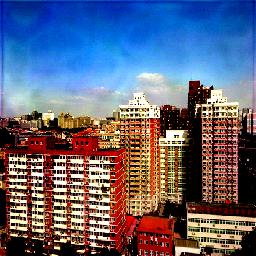}
\includegraphics[width=2.2cm,height=1.5cm]{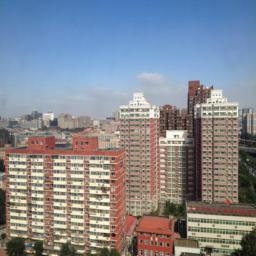}
\vspace{2pt}

\includegraphics[width=2.2cm,height=1.5cm]{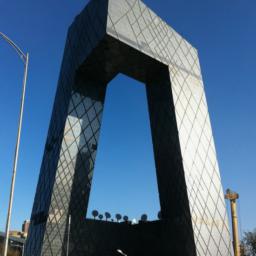}
\includegraphics[width=2.2cm,height=1.5cm]{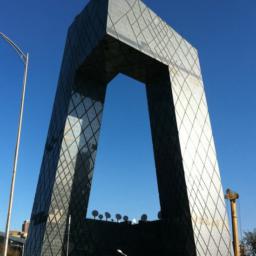}
\includegraphics[width=2.2cm,height=1.5cm]{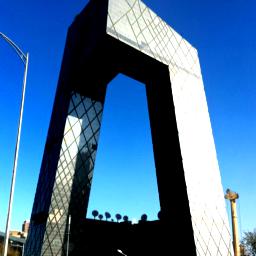}
\includegraphics[width=2.2cm,height=1.5cm]{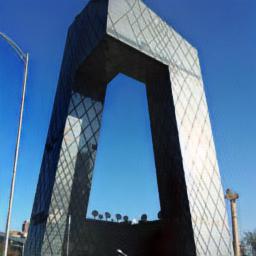}
\includegraphics[width=2.2cm,height=1.5cm]{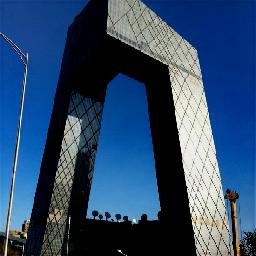}
\includegraphics[width=2.2cm,height=1.5cm]{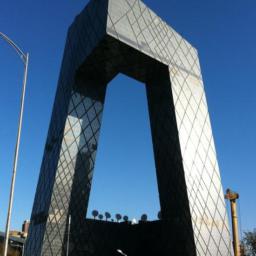}
\\
\leftline{\hspace{1.0cm} (g) GridDehazeNet \hspace{0.1cm}  (h) MSBDN  \hspace{0.7cm}   (i) SSID  \hspace{0.7cm}  (j) Cycle-Dehaze  \hspace{0.7cm} (k) ZID  \hspace{0.9cm}  (l) clear  \hspace{0.5cm}}
\caption{Visual results on RESIDE SOTS outdoor.}
\label{fig:vis_outdoor}
\end{figure}

Figure \ref{fig:vis_indoor} and Figure \ref{fig:vis_outdoor} show the visual results achieved by the supervised, semi-supervised and unsupervised algorithms, respectively, where the loss function used by the supervised algorithms is the L1 loss. The visual results show that the dehazed images obtained by the supervised algorithms are closer to the real images in terms of color and detail than the semi-supervised/unsupervised algorithms.

\begin{table}
\centering
\caption{Results on RESIDE SOTS indoor, where the SSIM and PNSR values are separated by the slash.}
\label{tab:results_RESIDE_indoor}
\begin{tabular}{ccccc}
\hline
Config & L1 & L1 + P & L2  & L2 + P \\
\hline
AODNet & 0.839/19.375 & 0.841/19.454 & 0.851/19.576 & 0.839/19.388 \\
4kDehazing & 0.932/23.881 & 0.949/26.569 & 0.928/23.370 & 0.928/23.353 \\
DMMDF & 0.959/30.585 & 0.960/30.383 & 0.961/30.631 &  0.961/30.725 \\
FFA-Net & \textbf{0.983}/\textbf{32.730} & \textbf{0.983}/\textbf{32.536} & \textbf{0.978}/\textbf{32.224} & \textbf{0.978}/\textbf{32.178} \\
GCA-Net & 0.924/25.044 &  0.930/25.638 & 0.917/24.716 & 0.909/24.714 \\
GridDehazeNet & 0.962/25.671 & 0.962/25.655 & 0.935/22.940 & 0.940/23.052 \\
MSBDN & 0.955/28.341 & 0.955/28.562 & 0.941/27.237 & 0.937/25.662 \\
\hline
SSID &  \multicolumn{4}{c}{0.814/20.959}  \\
Cycle-Dehaze & \multicolumn{4}{c}{0.810/18.880}\\
ZID & \multicolumn{4}{c}{0.835/19.830} \\
\hline
\end{tabular}
\end{table}

\begin{table}
\centering
\caption{Results on RESIDE SOTS outdoor, where the SSIM and PNSR values are separated by the slash.}
\label{tab:results_RESIDE_outdoor}
\begin{tabular}{ccccccccc}
\hline
Config & L1 & L1 + P & L2  & L2 + P \\
\hline
AODNet & 0.913/23.613 & 0.917/23.683 & 0.912/23.253 & 0.916/23.488  \\
4kDehazing & 0.963/28.476 & \textbf{0.964}/28.473 & 0.958/28.103 & 0.950/27.546 \\
DMMDF & 0.905/25.805 & 0.963/30.237 & 0.963/\textbf{30.535} & \textbf{0.965}/\textbf{30.682} \\
FFA-Net & 0.921/27.126 & 0.925/27.299 & 0.913/28.176 & 0.920/28.441 \\
GCA-Net & 0.953/27.784 & 0.949/27.559 & 0.946/27.20 & 0.943/27.273 \\
GridDehazeNet & \textbf{0.964}/28.296 & \textbf{0.964}/28.388 & 0.963/27.807 & 0.963/27.766 \\
MSBDN & 0.962/\textbf{29.944} & 0.963/\textbf{30.277} & \textbf{0.964}/29.843 & 0.956/28.782 \\
\hline
SSID & \multicolumn{4}{c}{0.840/20.905}  \\
Cycle-Dehaze & \multicolumn{4}{c}{0.861/20.347} \\
ZID & \multicolumn{4}{c}{0.633/13.520} \\
\hline
\end{tabular}
\end{table}

\begin{figure}
\centering
\includegraphics[width=9.6cm,height=7.7cm]{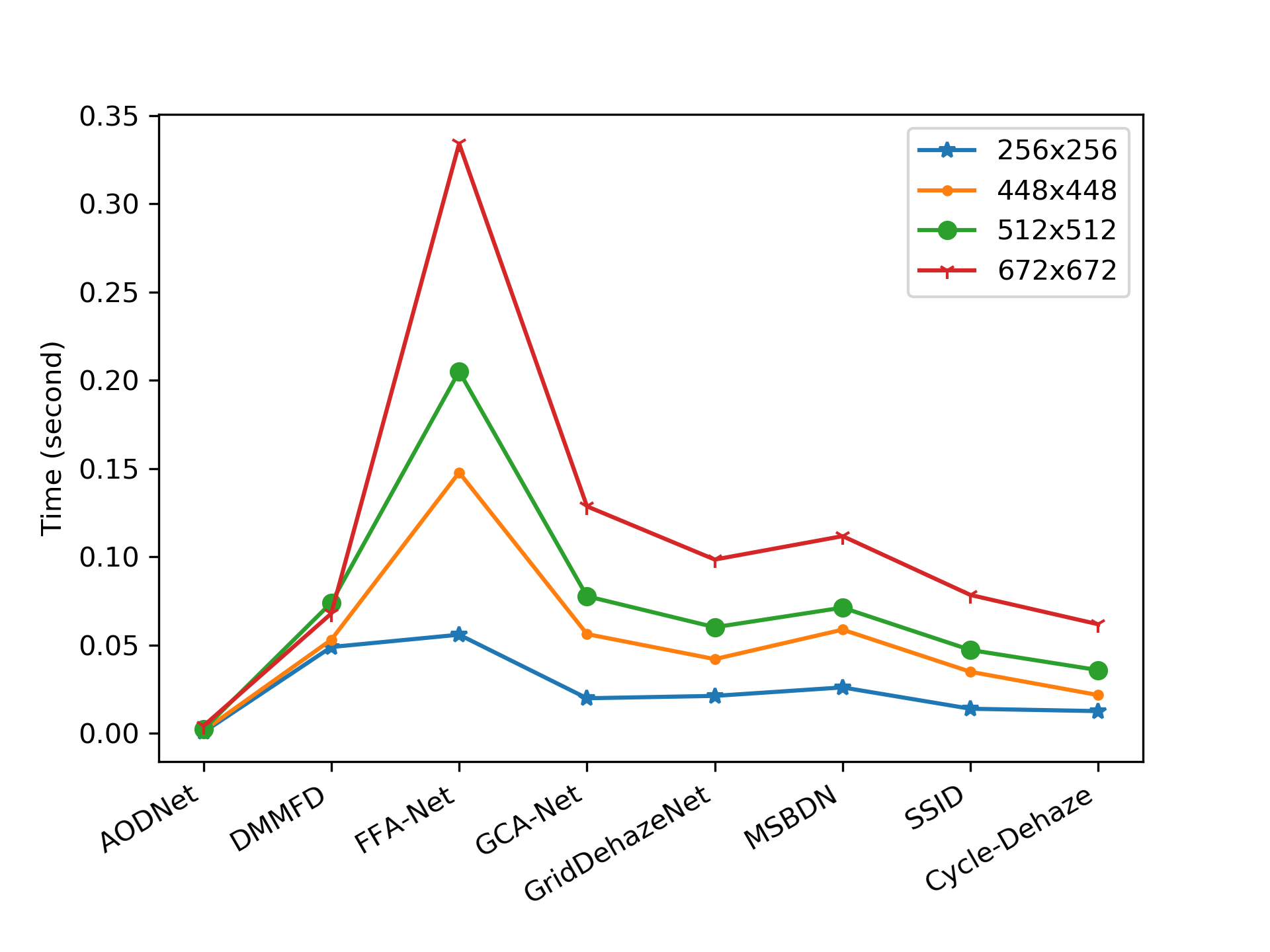}
\caption{Running speed on different sizes of input.}
\label{fig:speed}
\end{figure}

For a fair comparison of the computational speed of the baseline methods, the zero-shot-based ZID and high-resolution-based 4kDehazing are excluded. Fig. \ref{fig:speed} shows the average time for each algorithm that run 1000 times with PyTorch framework. It can be seen that AODNet is the fastest, and can achieve real-time effects for inputs of different sizes. FFA-Net is the slowest, taking more than 0.3 seconds to process the $672 \times 672$ input.

\section{Challenges and Opportunities}
\label{sec:future}
For image dehazing task, the current supervised, semi-supervised and unsupervised methods have achieved good performance. However, there are problems that still exist and open issues that should be explored in the future research. Next, we will discuss these challenges.

\subsection{More Effective ASM}
The current ASM has been proved to be suitable for describing the formation process of haze by many supervised, semi-supervised and unsupervised dehazing methods. However, recent work~\cite{ju2021ide} finds that the intrinsic limitation of ASM will cause a dim effect in the image after dehazing. By adding a new parameter, an enhanced ASM (EASM)~\cite{ju2021ide} is proposed. Improvements to ASM will have an important impact on the dehazing performance of existing methods that rely on ASM. Therefore, it is worth exploring a more accurate model of the haze formation process.

\subsection{Shift between the Real Domain and Synthetic Domain}
The current training process of dehazing models generally requires sufficient data. Since it is difficult to collect pairs of hazy and haze-free images in the real world, the synthesized data is needed by dehazing task. However, there are inherent differences between the synthesized hazy image and the real world hazy image. In order to solve the domain shift problem, the following three directions are worth exploring.
\begin{itemize}
\item The haze synthesized based on ASM cannot completely simulate the formation of haze in the real world, so we can attempt to design a more realistic hazy image synthesis algorithm to compensate for the difference between domains.
\item By introducing domain adaptation, semi-supervised and unsupervised methods into network design, experiments show that these algorithms can perform well in real world dehazing task.
\item Recent work (MRFID/BeDDE) collected some real world paired data, but they did not contain as many examples as RESIDE. It is challenging to build a large-scale real world dataset that can be used for a large-capacity CNN-based dehazing model.
\end{itemize}

\subsection{Computational Efficiency and New Metrics}
The dehazing models are often used as a preprocessing module for high-level computer vision tasks. For example, the lightweight AOD-Net is applied to object detection task. There are three issues that should be considered in the application of the dehazing model.
\begin{itemize}
\item In order to help follow-up tasks, the dehazing model should ensure that the dehazed image is of high quality.
\item The inference speed of the model must be fast enough to meet the real-time requirement.
\item Several end devices with small storage are sensitive to the number of parameters of the dehazing model.
\end{itemize}

Therefore, we need to balance quantitative performance, inference time and the number of parameters. High-quality dehazed image with high PSNR and SSIM are the main goals of the current research. FAMED-Net~\cite{zhang2020famed} discusses the computational efficiency of 14 models, which shows that several algorithms with excellent performance may be very time and memory consuming. Dehazing algorithms based on deep learning usually require the use of graphical computing units for model training and deployment. In real world applications, the forward inference speed of an algorithm is an important evaluation metric. 4kDehazing~\cite{zheng2021ultra} explores fast dehazing of high-resolution input, which takes only $8$ ms ($125$ fps) to process a 4K ($3840 \times 2160$) image on a single Titan RTX GPU. Future research can try to design a new evaluation metric that can comprehensively consider the dehazing quality, running speed and model size.

\subsection{Perceptual Loss}
The pre-trained model trained on a large-scale dataset can be used to calculate the perceptual loss. As shown in Table \ref{tab:loss_functions}, many dehazing models use the perceptual loss obtained by VGG16 or VGG19 as a part of the overall loss to improve the quality of the final result. Thus, a natural question is, can the perceptual loss be the same as the L1/L2 loss as a general loss for image dehazing task? Alternatively, is there a more efficient and higher quality way to compute the perceptual loss? For example, is it a better solution to obtain perceptual loss by other pre-trained models? Currently, there is no comprehensive study on the relationship between perceptual loss and dehazing task.

\subsection{How Dehazing Methods Affect High-level Computer Vision Tasks}
Many dehazing algorithms have been validated on high-level computer vision tasks, such as image segmentation and object detection. Experiments show that these dehazing methods can promote the performance of high-level computer vision tasks under hazy conditions. However, the recent study~\cite{pei2018does} has shown that several dehazing algorithms may be ineffective for image classification, although their dehazing ability has been proven. The experimental results~\cite{pei2018does} on synthetic and real hazy data show that several well-designed dehazing models have little positive effect on the performance of the classification model, and sometimes may reduce the classification accuracy to some extent.

Besides, domain adaptation and generalization methods have been proposed to deal with high-level computer vision tasks such as object detection \cite{wang2022robust,wang2021exploring} and semantic segmentation \cite{zhang2019category,gao2021dsp} in bad weather conditions including haze, which can be treated as mitigating the side effect of haze in the feature space by aligning hazy image features with those clean ones. It will be an interesting research topic to investigate and reduce the influence of haze on high-level computer vision tasks both at image-level (i.e., dehazing) and feature-level (i.e., domain adaptation).

\subsection{Prior Knowledge and Learning Model}
Before deep learning was widely used in image dehazing task, image-based prior statistical knowledge was an important part for guiding the dehazing process. Now, extensive work has shown that deep learning techniques can effectively remove haze from images independently of physical model. At the same time, several recent studies have found that prior statistical knowledge can be used for semi-supervised~\cite{Li2019semi} and unsupervised~\cite{alona2020un_dcp} network training. Effective prior knowledge can reduce the model's dependence on data to a certain extent and improve the generalization ability of the model. However, there is currently no evidence which statistical priors are always valid. Therefore, the general prior knowledge that can be used for the CNN-based dehazing algorithm is worth verifying.

\section{Conclusion}
This paper provides a comprehensive survey of deep learning-based dehazing research. First, the commonly used physical model, high-quality datasets, general loss functions, effective network modules and evaluation metrics are summarized. Then, supervised, semi-supervised and unsupervised dehazing studies are classified and analyzed from different technical perspectives. Moreover, quantitative and qualitative dehazing performance of various baselines are discussed. Finally, we discuss several valuable research directions and open issues.



\begin{acks}
This work was supported in part by the grant of the National Science Foundation of China under Grant 62172090, 62172089; Alibaba Group through Alibaba Innovative Research Program; CAAI-Huawei MindSpore Open Fund. All correspondence should be directed to Xiaofeng Cong and Yuan Cao. Dr Jing Zhang is supported by ARC research project FL-170100117.
\end{acks}

\appendix
\section{What is not discussed in this survey}
We noticed that several papers' formulation description and open source code may not be consistent. In particular, loss functions not provided in several papers are used in the training code. In this case, we use the loss functions provided in the paper as a guide, regardless of the code implementation.
Further, several public papers that have not yet been published formally are not covered in this survey, such as those on arxiv, since their content may be changed in the future.
\bibliographystyle{ACM-Reference-Format}
\bibliography{refs}


\end{document}